\documentclass[11pt,a4paper]{article}
\usepackage[hyperref]{acl2019}
\usepackage{times}
\usepackage{latexsym}

\usepackage[T1]{fontenc}  
\usepackage[utf8]{inputenc}
\usepackage[english]{babel} 

\usepackage{longtable}
\usepackage{caption}

\usepackage{placeins}

\usepackage{url}
\usepackage{changepage}

\usepackage{hyperref}
\usepackage{url}
\usepackage{graphicx}
\usepackage{multirow}
\usepackage{booktabs}
\usepackage[inline]{enumitem}
\usepackage{caption}
\usepackage{subcaption}
\usepackage{xcolor}
\usepackage{nccmath}

\usepackage{amsmath,amsfonts,bm}

\def\eqref#1{equation~\ref{#1}}

\def\1{\bm{1}}

\def\rva{{\mathbf{a}}}

\def\rvd{{\mathbf{d}}}

\def\rvh{{\mathbf{h}}}

\def\rvk{{\mathbf{k}}}

\def\rvm{{\mathbf{m}}}

\def\rvo{{\mathbf{o}}}

\def\rvq{{\mathbf{q}}}

\def\rvs{{\mathbf{s}}}

\def\rvv{{\mathbf{v}}}

\def\rvx{{\mathbf{x}}}

\def\rmA{{\mathbf{A}}}
\def\rmB{{\mathbf{B}}}

\def\rmD{{\mathbf{D}}}
\def\rmE{{\mathbf{E}}}

\def\rmQ{{\mathbf{Q}}}
\def\rmR{{\mathbf{R}}}

\def\rmU{{\mathbf{U}}}

\def\rmW{{\mathbf{W}}}

\DeclareMathAlphabet{\mathsfit}{\encodingdefault}{\sfdefault}{m}{sl}
\SetMathAlphabet{\mathsfit}{bold}{\encodingdefault}{\sfdefault}{bx}{n}

\newcommand{\R}{\mathbb{R}}

\newcommand{\rulesep}{\unskip\ \vrule\ }
\usepackage{amssymb}
\usepackage{pifont}
\newcommand{\cmark}{\ding{51}}%
\newcommand{\xmark}{\ding{55}}%

\usepackage{mathtools}

\newcommand{\red}[1]{{\color{red}#1}}

\newcommand{\cyan}[1]{{\color{cyan}#1}}

\newcommand{\magenta}[1]{{\color{magenta}#1}}

\aclfinalcopy 

\title{Transformer-XL: Attentive Language Models \\ Beyond a Fixed-Length Context}

\author{Zihang Dai$^{*12}$, Zhilin Yang$^{*12}$, Yiming Yang$^1$, Jaime Carbonell$^1$, \\
	{\bf Quoc V. Le$^2$, Ruslan Salakhutdinov$^1$ }\\
	$^1$Carnegie Mellon University, $^2$Google Brain \\
	{\small \texttt{\{dzihang,zhiliny,yiming,jgc,rsalakhu\}@cs.cmu.edu, qvl@google.com} } 
}

\date{}

\begin{document}
\maketitle

\renewcommand{\thefootnote}{\fnsymbol{footnote}}
\footnotetext[1]{Equal contribution. Order determined by swapping the one in \citet{yang2017breaking}.}
\renewcommand{\thefootnote}{\arabic{footnote}}

\begin{abstract}

Transformers have a potential of learning longer-term dependency, but are limited by a fixed-length context in the setting of language modeling.
We propose a novel neural architecture \textit{Transformer-XL} that enables learning dependency beyond a fixed length without disrupting temporal coherence. It consists of a segment-level recurrence mechanism and a novel positional encoding scheme. Our method not only enables capturing longer-term dependency, but also resolves the context fragmentation problem. As a result, Transformer-XL learns dependency that is 80\% longer than RNNs and 450\% longer than vanilla Transformers, achieves better performance on both short and long sequences, and is up to 1,800+ times faster than vanilla Transformers during evaluation.
Notably, we improve the state-of-the-art results of bpc/perplexity to 0.99 on enwiki8, 1.08 on text8, 18.3 on WikiText-103, 21.8 on One Billion Word, and 54.5 on Penn Treebank (without finetuning).
When trained only on WikiText-103, Transformer-XL manages to generate reasonably coherent, novel text articles with thousands of tokens.
Our code, pretrained models, and hyperparameters are available in both Tensorflow and PyTorch\footnote{\url{https://github.com/kimiyoung/transformer-xl}}.

\end{abstract}
\section{Introduction}
\label{sec:intro}

Language modeling is among the important problems that require modeling long-term dependency, with successful applications such as unsupervised pretraining~\citep{dai2015semi,peters2018deep,radford2018improving,devlin2018bert}.
However, it has been a challenge to equip neural networks with the capability to model long-term dependency in sequential data.
Recurrent neural networks (RNNs), in particular Long Short-Term Memory (LSTM) networks~\citep{hochreiter1997long}, have been a standard solution to language modeling and obtained strong results on multiple benchmarks.
Despite the wide adaption, RNNs are difficult to optimize due to gradient vanishing and explosion~\citep{hochreiter2001gradient}, and the introduction of gating in LSTMs and the gradient clipping technique~\citep{graves2013generating} might not be sufficient to fully address this issue.
Empirically, previous work has found that LSTM language models use 200 context words on average~\citep{khandelwal2018sharp}, indicating room for further improvement.

On the other hand, the direct connections between long-distance word pairs baked in attention mechanisms might ease optimization and enable the learning of long-term dependency~\citep{bahdanau2014neural,vaswani2017attention}.
Recently, \citet{al2018character} designed a set of auxiliary losses to train deep Transformer networks for character-level language modeling, which outperform LSTMs by a large margin.
Despite the success, the LM training in~\citet{al2018character} is performed on separated fixed-length segments of a few hundred characters, without any information flow across segments.
As a consequence of the fixed context length, the model cannot capture any longer-term dependency beyond the predefined context length.
In addition, the fixed-length segments are created by selecting a consecutive chunk of symbols without respecting the sentence or any other semantic boundary.
Hence, the model lacks necessary contextual information needed to well predict the first few symbols, leading to inefficient optimization and inferior performance.
We refer to this problem as \textit{context fragmentation}.



To address the aforementioned limitations of fixed-length contexts, we propose a new architecture called Transformer-XL (meaning extra long).
We introduce the notion of recurrence into our deep self-attention network. In particular, instead of computing the hidden states from scratch for each new segment, we reuse the hidden states obtained in previous segments.
The reused hidden states serve as memory for the current segment, which builds up a recurrent connection between the segments.
As a result, modeling very long-term dependency becomes possible because information can be propagated through the recurrent connections.
Meanwhile, passing information from the previous segment can also resolve the problem of context fragmentation.
More importantly, we show the necessity of using relative positional encodings rather than absolute ones, in order to enable state reuse without causing temporal confusion.
Hence, as an additional technical contribution, we introduce a simple but more effective relative positional encoding formulation that generalizes to attention lengths longer than the one observed during training.

Transformer-XL obtained strong results on five datasets, varying from word-level to character-level language modeling.
Transformer-XL is also able to generate relatively coherent long text articles with \textit{thousands of} tokens (see Appendix \ref{sec:gen}), trained on only 100M tokens.

Our main technical contributions include introducing the notion of recurrence in a purely self-attentive model and deriving a novel positional encoding scheme. These two techniques form a complete set of solutions, as any one of them alone does not address the issue of fixed-length contexts. Transformer-XL is the first self-attention model that achieves substantially better results than RNNs on both character-level and word-level language modeling.

\section{Related Work}
In the last few years, the field of language modeling has witnessed many significant advances, including but not limited to devising novel architectures to better encode the context~\citep{bengio2003neural,mikolov2010recurrent,
merity2016pointer,al2018character}, improving regularization and optimization algorithms~\cite{gal2016theoretically}
, speeding up the Softmax computation~\citep{grave2016efficient}
, and enriching the output distribution family~\citep{yang2017breaking}.

To capture the long-range context in language modeling, a line of work directly feeds a representation of the wider context into the network as an additional input.
Existing works range from ones where context representations are manually defined~\citep{mikolov2012context,ji2015document,wang2015larger} to others that rely on document-level topics learned from data~\citep{dieng2016topicrnn,wang2017topic}.

More broadly, in generic sequence modeling, how to capture long-term dependency has been a long-standing research problem.
From this perspective, since the ubiquitous adaption of LSTM, many efforts have been spent on relieving the vanishing gradient problem, including better initialization~\citep{le2015simple}, additional loss signal~\citep{trinh2018learning}, augmented memory structure~\citep{ke2018sparse} and others that modify the internal architecture of RNNs to ease the optimization~\cite{wu2016multiplicative,li2018independently}.
Different from them, our work is based on the Transformer architecture and shows that language modeling as a real-world task benefits from the ability to learn longer-term dependency.

\section{Model} \label{sec:model}
\begin{figure*}[t]
	\begin{subfigure}[b]{0.292\linewidth}
		\includegraphics[width=\textwidth]{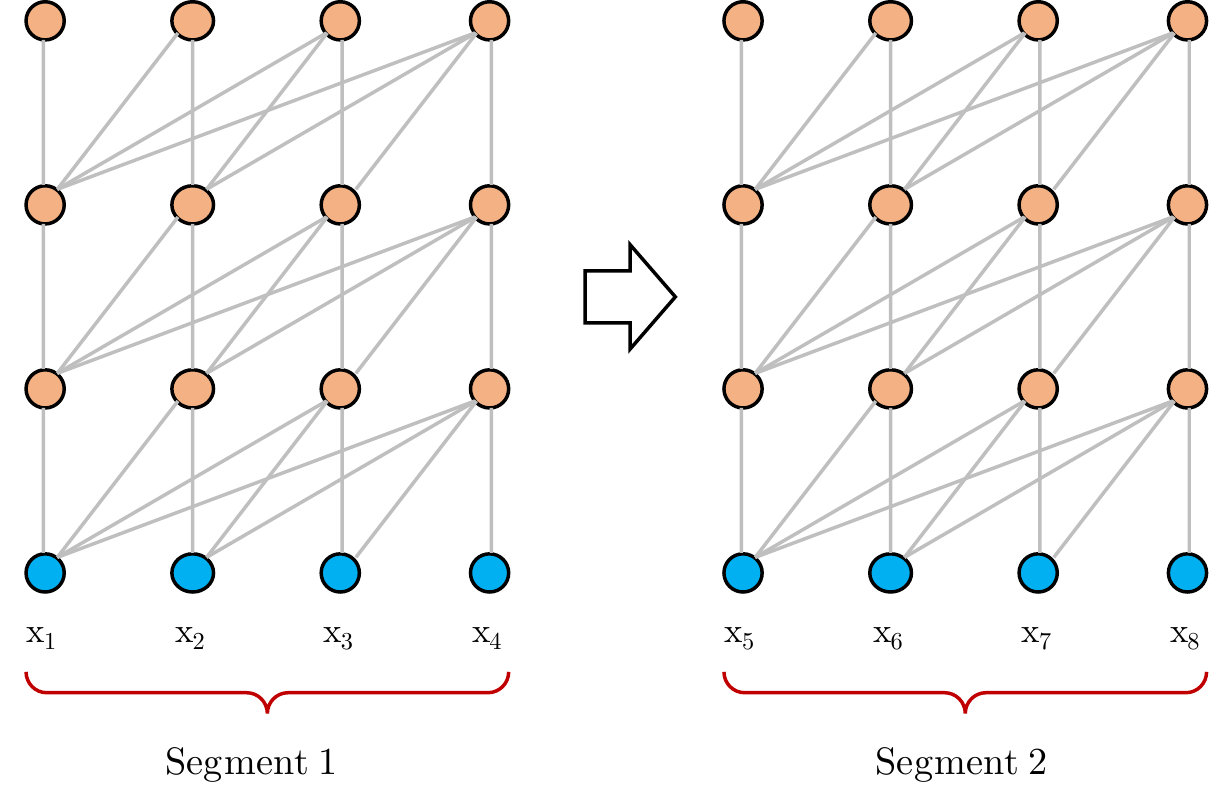}
		\caption{\small Train phase.}
		\label{fig:vanilla-train}
	\end{subfigure}
	\rulesep
	\begin{subfigure}[b]{0.69\linewidth}
		\includegraphics[width=\textwidth]{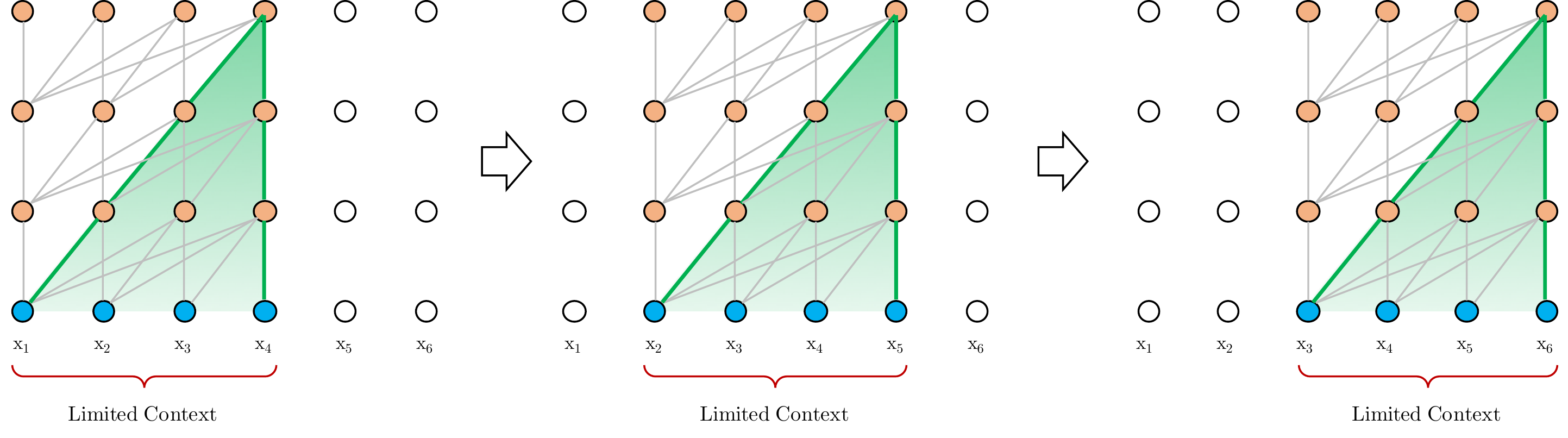}
		\caption{\small Evaluation phase.}
		\label{fig:vanilla-eval}
	\end{subfigure}
	\caption{\small Illustration of the vanilla model with a segment length 4.}
	\label{fig:vanilla}
\vspace{-1em}
\end{figure*}

Given a corpus of tokens $\rvx = (x_1 , \dots, x_T)$, the task of language modeling is to estimate the joint probability $P(\rvx)$, which is often auto-regressively factorized as $P(\rvx) = \prod_{t} P(x_t \mid \rvx_{<t})$.
With the factorization, the problem reduces to estimating each conditional factor.
In this work, we stick to the standard neural approach to modeling the conditional probability.
Specifically, a trainable neural network is used to encode the context $\rvx_{<t}$ into a fixed size hidden state, which is multiplied with the word embeddings to obtain the logits.
The logits are then fed into the Softmax function, yielding a categorical probability distribution over the next token.

\subsection{Vanilla Transformer Language Models}
In order to apply Transformer or self-attention to language modeling, the central problem is how to train a Transformer to effectively encode an arbitrarily long context into a fixed size representation.
Given infinite memory and computation, a simple solution would be to process the entire context sequence using an unconditional Transformer decoder, similar to a feed-forward neural network.
However, this is usually infeasible with the limited resource in practice.

One feasible but crude approximation is to split the entire corpus into shorter segments of manageable sizes, and only train the model within each segment, ignoring all contextual information from previous segments.
This is the idea adopted by \citet{al2018character}. We call it the \textit{vanilla model} and visualize it in Fig. \ref{fig:vanilla-train}.
Under this training paradigm, information never flows across segments in either the forward or backward pass.
There are two critical limitations of using a fixed-length context.
First, the largest possible dependency length is upper bounded by the segment length, which is a few hundred on character-level language modeling \citep{al2018character}.
Therefore, although the self-attention mechanism is less affected by the vanishing gradient problem compared to RNNs, the vanilla model is not able to fully exploit this optimization advantage.
Second, though it is possible to use padding to respect the sentence or other semantic boundaries, in practice it has been standard practice to simply chunk long text into fixed-length segments due to improved efficiency \citep{peters2018deep,devlin2018bert,al2018character}. However, simply chunking a sequence into fixed-length segments will lead to the context fragmentation problem as discussed in Section \ref{sec:intro}.

During evaluation, at each step, the vanilla model also consumes a segment of the same length as in training, but only makes one prediction at the last position.
Then, at the next step, the segment is shifted to the right by only one position, and the new segment has to be processed all from scratch.
As shown in Fig.~\ref{fig:vanilla-eval}, this procedure ensures that each prediction utilizes the longest possible context exposed during training, and also relieves context fragmentation issue encountered in training. However, this evaluation procedure is extremely expensive. We will show that our proposed architecture is able to substantially improve the evaluation speed.

\subsection{Segment-Level Recurrence with State Reuse}



\begin{figure*}[!h]
	\begin{subfigure}[b]{0.62\linewidth}
		\includegraphics[width=\textwidth]{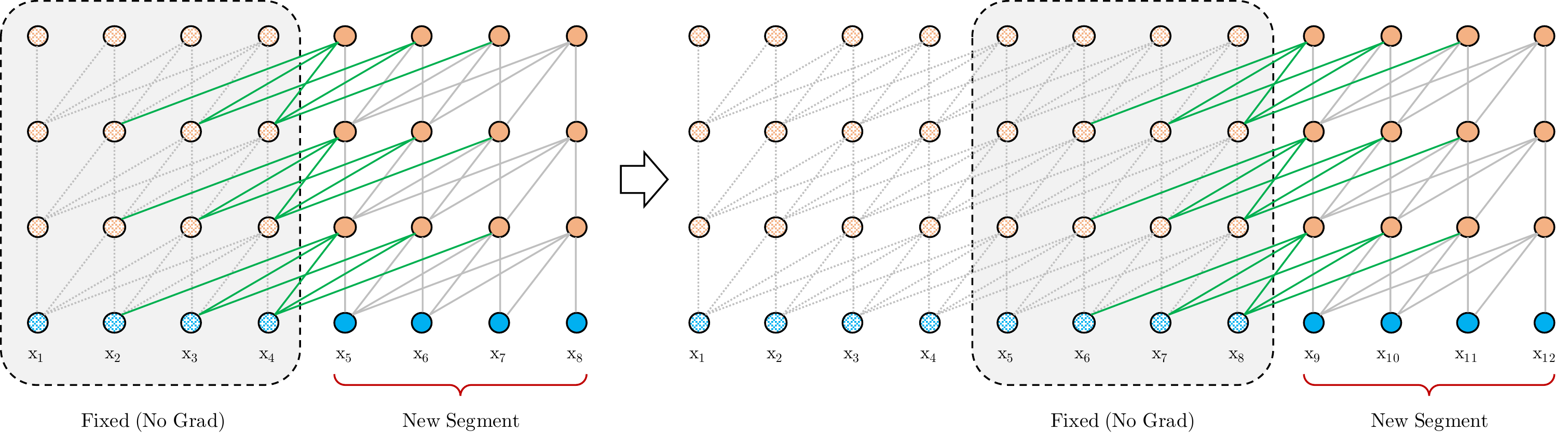}
		\caption{\small Training phase.}
		\label{fig:xl-train}
	\end{subfigure}
	\rulesep
	\begin{subfigure}[b]{0.35\linewidth}
		\includegraphics[width=\textwidth]{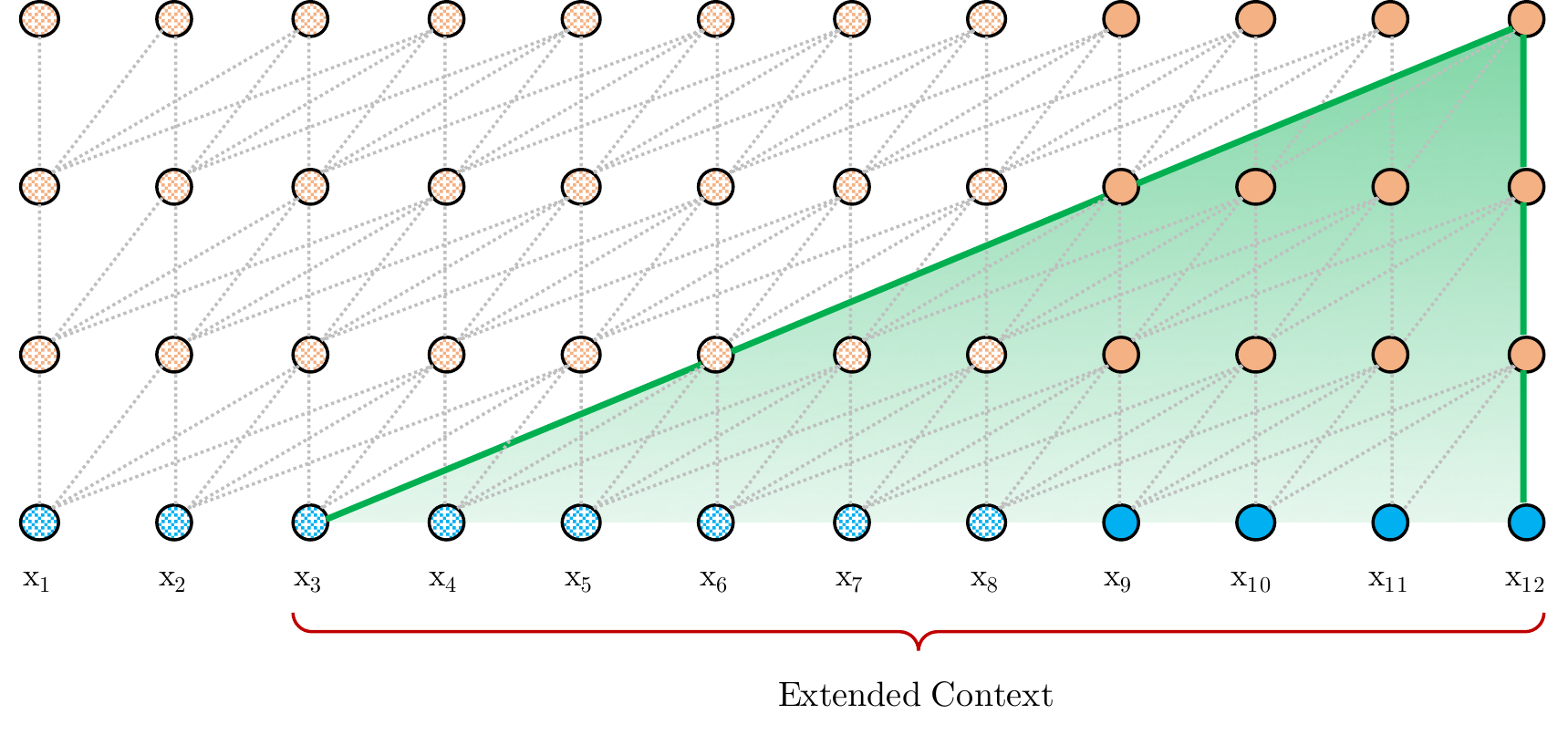}
		\caption{\small Evaluation phase.}
		\label{fig:xl-eval}
	\end{subfigure}
	\caption{\small Illustration of the Transformer-XL model with a segment length 4.}
	\label{fig:xl}
\vspace{-1em}
\end{figure*}
To address the limitations of using a fixed-length context, we propose to introduce a recurrence mechanism to the Transformer architecture.
During training, the hidden state sequence computed for the previous segment is \textit{fixed} and \textit{cached} to be reused as an extended context when the model processes the next new segment, as shown in Fig. \ref{fig:xl-train}.
Although the gradient still remains within a segment, this additional input allows the network to exploit information in the history, leading to an ability of modeling longer-term dependency and avoiding context fragmentation.
Formally, let the two consecutive segments of length $L$ be $\rvs_{\tau} = \left[x_{\tau,1}, \cdots, x_{\tau,L}\right]$ and $\rvs_{\tau+1} = \left[x_{\tau+1,1}, \cdots, x_{\tau+1,L}\right]$ respectively.
Denoting the $n$-th layer hidden state sequence produced for the $\tau$-th segment $\rvs_{\tau}$ by $\rvh_{\tau}^{n} \in \R^{L \times d}$, where $d$ is the hidden dimension.
Then, the $n$-th layer hidden state for segment $\rvs_{\tau+1}$ is produced (schematically) as follows,
\par\nobreak
\vspace{-0.5em}
\small
\begin{align*}\label{eqn:reuse}
	&\widetilde{\rvh}_{\tau+1}^{n-1} = \left[ \text{SG}(\rvh_{\tau}^{n-1}) \circ \rvh_{\tau+1}^{n-1} \right],
		\\
	&\rvq_{\tau+1}^{n}, \rvk_{\tau+1}^{n}, \rvv_{\tau+1}^{n} = \rvh_{\tau+1}^{n-1} \rmW_q^\top, \widetilde{\rvh}_{\tau+1}^{n-1} \rmW_k^\top, \widetilde{\rvh}_{\tau+1}^{n-1} \rmW_v^\top,
		\\
	&\rvh_{\tau+1}^{n} = \text{Transformer-Layer}\left(\rvq_{\tau+1}^{n}, \rvk_{\tau+1}^{n}, \rvv_{\tau+1}^{n}\right).
\end{align*}
\normalsize
\vspace{-1.5em}

\noindent where the function $\text{SG}(\cdot)$ stands for stop-gradient, the notation $[\rvh_u \circ \rvh_v]$ indicates the concatenation of two hidden sequences along the length dimension, and $\mathbf{W}_\cdot$ denotes model parameters.
Compared to the standard Transformer, the critical difference lies in that the key $\rvk_{\tau+1}^{n}$ and value $\rvv_{\tau+1}^{n}$ are conditioned on the extended context $\widetilde{\rvh}_{\tau+1}^{n-1}$ and hence $\rvh_{\tau}^{n-1}$ cached from the previous segment.
We emphasize this particular design by the green paths in Fig. \ref{fig:xl-train}.

With this recurrence mechanism applied to every two consecutive segments of a corpus, it essentially creates a segment-level recurrence in the hidden states.
As a result, the effective context being utilized can go way beyond just two segments.
However, notice that the recurrent dependency between $\rvh_{\tau+1}^{n}$ and $\rvh_{\tau}^{n-1}$ shifts one layer downwards per-segment, which differs from the same-layer recurrence in conventional RNN-LMs.
Consequently, the largest possible dependency length grows linearly w.r.t. the number of layers as well as the segment length, i.e., $O(N \times L)$, as visualized by the shaded area in Fig. \ref{fig:xl-eval}.
This is analogous to truncated BPTT \citep{mikolov2010recurrent}, a technique developed for training RNN-LMs. However, different from truncated BPTT, our method caches a sequence of hidden states instead of the last one, and should be applied together with the relative positional encoding technique described in Section \ref{sec:rel-pos-embed}.

Besides achieving extra long context and resolving fragmentation, another benefit that comes with the recurrence scheme is significantly faster evaluation.
Specifically, during evaluation, the representations from the previous segments can be reused instead of being computed from scratch as in the case of the vanilla model.
In our experiments on enwiki8, Transformer-XL is up to 1,800+ times faster than the vanilla model during evaluation (see Section \ref{sec:exp}).


Finally, notice that the recurrence scheme does not need to be restricted to only the previous segment.
In theory, we can cache as many previous segments as the GPU memory allows, and reuse all of them as the extra context when processing the current segment.
Thus, we can cache a predefined length-$M$ old hidden states spanning (possibly) multiple segments, and refer to them as the memory $\rvm_{\tau}^{n} \in \R^{M \times d}$, due to a clear connection to the memory augmented neural networks~\citep{graves2014neural,weston2014memory}.
In our experiments, we set $M$ equal to the segment length during training, and increase it by multiple times during evaluation.

\subsection{Relative Positional Encodings}
\label{sec:rel-pos-embed}
While we found the idea presented in the previous subsection very appealing, there is a crucial technical challenge we haven't solved in order to reuse the hidden states.
That is, how can we keep the positional information coherent when we reuse the states?
Recall that, in the standard Transformer, the information of sequence order is provided by a set of positional encodings, denoted as $\rmU \in \R^{L_\text{max} \times d}$, where the $i$-th row $\rmU_i$ corresponds to the $i$-th \textit{absolute} position within a segment and $L_\text{max}$ prescribes the maximum possible length to be modeled.
Then, the actual input to the Transformer is the element-wise addition of the word embeddings and the positional encodings.
If we simply adapt this positional encoding to our recurrence mechanism, the hidden state sequence would be computed schematically by
\par\nobreak
\vspace{-0.5em}
\small
\begin{align*}
	\rvh_{\tau+1} &= f(\rvh_{\tau}, \rmE_{\rvs_{\tau+1}} + \rmU_{1:L}) \\
	\rvh_{\tau} &= f(\rvh_{\tau-1}, \rmE_{\rvs_{\tau}} + \rmU_{1:L}),
\end{align*}
\normalsize
\vspace{-1.5em}

\noindent where $\rmE_{\rvs_{\tau}} \in \R^{L \times d}$ is the word embedding sequence of $\rvs_{\tau}$, and $f$ represents a transformation function.
Notice that, both $\rmE_{\rvs_{\tau}}$ and $\rmE_{\rvs_{\tau+1}}$ are associated with the same positional encoding $\rmU_{1:L}$.
As a result, the model has no information to distinguish the positional difference between $x_{\tau,j}$ and $x_{\tau+1,j}$ for any $j = 1, \dots, L$, resulting in a sheer performance loss.

In order to avoid this failure mode, the fundamental idea is to only encode the \textit{relative} positional information in the hidden states.
Conceptually, the positional encoding gives the model a temporal clue or ``bias'' about how information should be gathered, i.e., where to attend.
For the same purpose, instead of incorporating bias statically into the initial embedding, one can inject the same information into the attention score of each layer.
More importantly, it is more intuitive and generalizable to define the temporal bias in a relative manner.
For instance, when a query vector $q_{\tau,i}$ attends on the key vectors $\rvk_{\tau,\leq i}$, it does not need to know the absolute position of each key vector to identify the temporal order of the segment.
Instead, it suffices to know the relative distance between each key vector $k_{\tau,j}$ and itself $q_{\tau,i}$, i.e. $i - j$. 
Practically, one can create a set of relative positional encodings $\rmR \in \R^{L_\text{max} \times d}$, where the $i$-th row $\rmR_i$ indicates a relative distance of $i$ between two positions.
By injecting the relative distance dynamically into the attention score, the query vector can easily distinguish the representations of $x_{\tau,j}$ and $x_{\tau+1,j}$ from their different distances, making the state reuse mechanism feasible.
Meanwhile, we won't lose any temporal information, as the absolute position can be recovered recursively from relative distances.

Previously, the idea of relative positional encodings has been explored in the context of machine translation~\citep{shaw2018self} and music generation~\citep{huang2018improved}.
Here, we offer a different derivation, arriving at a new form of relative positional encodings, which not only has a one-to-one correspondence to its absolute counterpart but also enjoys much better generalization empirically (see Section \ref{sec:exp}).
Firstly, in the standard Transformer \citep{vaswani2017attention}, the attention score between query $q_i$ and key vector $k_j$ within the same segment can be decomposed as
\par\nobreak
\vspace{-0.5em}
\small
\begin{align*}
	\rmA_{i,j}^\text{abs} 
	&= \underbrace{ \rmE_{x_i}^\top \rmW_q^\top \rmW_k \rmE_{x_j} }_{(a)}
	+ \underbrace{\rmE_{x_i}^\top \rmW_q^\top \rmW_k \rmU_{j}}_{(b)} \\
	&+ \underbrace{\rmU_{i}^\top \rmW_q^\top \rmW_k \rmE_{x_j}}_{(c)}
	+ \underbrace{\rmU_{i}^\top \rmW_q^\top \rmW_k \rmU_{j}}_{(d)}. \vspace{-0.5em}
\end{align*}
\normalsize
\vspace{-1em}

Following the idea of only relying on relative positional information, we propose to re-parameterize the four terms as follows
\par\nobreak
\vspace{-0.5em}
\small
\begin{align*}
	\rmA_{i,j}^\text{rel}
	&= \underbrace{ \rmE_{x_i}^\top \rmW_q^\top \rmW_{k,E} \rmE_{x_j} }_{(a)}
	+ \underbrace{\rmE_{x_i}^\top \rmW_q^\top \rmW_{k,R} \cyan{\rmR_{i-j}} }_{(b)} \\
	&+ \underbrace{\red{u^\top} \rmW_{k,E} \rmE_{x_j}}_{(c)}
	+ \underbrace{\red{v^\top} \rmW_{k,R} \cyan{\rmR_{i-j}}}_{(d)}.
\end{align*}
\normalsize
\vspace{-1em}

\begin{itemize}[leftmargin=*,itemsep=0pt,parsep=0.5em,topsep=0pt,partopsep=0pt]
	\item The first change we make is to replace all appearances of the absolute positional embedding $\rmU_{j}$ for computing key vectors in term $(b)$ and $(d)$ with its relative counterpart $\cyan{\rmR_{i-j}}$.
	This essentially reflects the prior that only the relative distance matters for where to attend.
	Note that $\cyan{\rmR}$ is a sinusoid encoding matrix~\citep{vaswani2017attention} without learnable parameters.
	\item Secondly, we introduce a trainable parameter $\red{u} \in \R^{d}$ to replace the query $\rmU_{i}^\top \rmW_q^\top$ in term $(c)$.
	In this case, since the query vector is the same for all query positions, it suggests that the attentive bias towards different words should remain the same regardless of the query position.
	With a similar reasoning, a trainable parameter $\red{v} \in \R^{d}$ is added to substitute $\rmU_{i}^\top \rmW_q^\top$ in term $(d)$.
	\item Finally, we deliberately separate the two weight matrices $\rmW_{k,E}$ and $\rmW_{k,R}$ for producing the content-based key vectors and location-based key vectors respectively.
\end{itemize}
Under the new parameterization, each term has an intuitive meaning: term $(a)$ represents content-based addressing, term $(b)$ captures a content-dependent positional bias, term $(c)$ governs a global content bias, and $(d)$ encodes a global positional bias.

In comparison, the formulation in \citet{shaw2018self} only has terms $(a)$ and $(b)$, dropping the two bias terms $(c)$ and $(d)$.
Moreover, \citet{shaw2018self} merge the multiplication $\rmW_k \rmR$ into a single trainable matrix $\hat{\rmR}$, which abandons the inductive bias built into the original sinusoid positional encoding~\citep{vaswani2017attention}.
In contrast, our relative positional embedding $\rmR$ adapts the sinusoid formulation.
As a benefit of the inductive bias, a model trained on a memory of some certain length can automatically generalize to a memory several times longer during evaluation.

Equipping the recurrence mechanism with our proposed relative positional embedding, we finally arrive at the Transformer-XL architecture.
For completeness, we summarize the computational procedure for a $N$-layer Transformer-XL with a single attention head here. For $n = 1, \dots, N$:
\par\nobreak
\vspace{-0.5em}
\small
\begingroup
\allowdisplaybreaks
\begin{align*}
\widetilde{\rvh}_{\tau}^{n-1}
	=&\, \left[ \text{SG}(\rvm_{\tau}^{n-1}) \circ \rvh_{\tau}^{n-1} \right] \\
\rvq_{\tau}^{n}, \rvk_{\tau}^{n}, \rvv_{\tau}^{n}
	=&\, \rvh_{\tau}^{n-1} {\rmW_q^{n}}^\top, \widetilde{\rvh}_{\tau}^{n-1} {\rmW_{k,E}^{n}}^\top, \widetilde{\rvh}_{\tau}^{n-1} {\rmW_v^{n}}^\top \\
\rmA_{\tau,i,j}^{n}
	=&\, {\rvq_{\tau,i}^{n}}^\top \rvk_{\tau,j}^{n} + {\rvq_{\tau,i}^{n}}^\top \rmW_{k,R}^{n} \rmR_{i-j} \\&\,+ u^\top \rvk_{\tau,j} + v^\top \rmW_{k,R}^{n} \rmR_{i-j} \\
\rva_{\tau}^{n} =&\, \text{Masked-Softmax}(\rmA_{\tau}^{n}) \rvv_{\tau}^{n} \\
\rvo_{\tau}^{n} =&\, \text{LayerNorm}(\text{Linear}(\rva_{\tau}^{n}) + \rvh_{\tau}^{n-1}) \\
\rvh_{\tau}^{n} =&\, \text{Positionwise-Feed-Forward}(\rvo_{\tau}^{n})
\end{align*}
\endgroup
\normalsize
\vspace{-1.5em}

\noindent with $\rvh_{\tau}^{0} \coloneqq \rmE_{\rvs_{\tau}}$ defined as the word embedding sequence.
In addition, it is worth mentioning that a naive way to compute $\rmA$ requires computing $\rmW_{k,R}^{n} \rmR_{i-j}$ for all pairs $(i,j)$, whose cost is quadratic w.r.t. the sequence length.
However, noticing that the value of $i-j$ only ranges from zero to the sequence length,
we show a simple computation procedure in Appendix \ref{sec:A-efficient-attention}, which reduces the cost to be linear w.r.t. the sequence length.

\section{Experiments}
\label{sec:exp}

\subsection{Main Results}


\bgroup
\setlength{\tabcolsep}{3pt}
\begin{table}[t]
	\small
	\centering
	
	\begin{adjustwidth}{-4pt}{}
	\begin{tabular}{l|cc}
		\toprule
		\bf Model & \bf \#Param &  \bf PPL \\
		\midrule
		\citet{grave2016improving} - LSTM & - & 48.7 \\
		\citet{bai2018empirical} - TCN & - & 45.2 \\
		\citet{dauphin2016language} - GCNN-8 & - & 44.9 \\
		\citet{grave2016improving} - LSTM + Neural cache & - & 40.8 \\
		\citet{dauphin2016language} - GCNN-14 & - & 37.2 \\
		\citet{merity2018analysis} - QRNN & 151M & 33.0 \\
		\citet{rae2018fast} - Hebbian + Cache & - & 29.9 \\
		Ours - Transformer-XL Standard & 151M & \textbf{24.0} \\
		\midrule
		\citet{baevski2018adaptive} - Adaptive Input$^\diamond$ & 247M & 20.5 \\
		Ours - Transformer-XL Large & 257M & \textbf{18.3} \\
		\bottomrule
	\end{tabular}
	\caption{\small
		Comparison with state-of-the-art results on WikiText-103. $^\diamond$ indicates contemporary work.
	}
	\label{table:103}
	\end{adjustwidth}
\end{table}
\egroup

\bgroup
\setlength{\tabcolsep}{3pt}
\begin{table}[t]
    \small
    \centering
    \begin{tabular}{l|cc}
        \toprule
        \bf Model & \bf \#Param & \bf bpc \\
        \midrule
        \citet{ha2016hypernetworks} - LN HyperNetworks & 27M & 1.34 \\
        \citet{chung2016hierarchical} - LN HM-LSTM & 35M & 1.32 \\
        \citet{zilly2016recurrent} - RHN & 46M & 1.27 \\
        \citet{mujika2017fast} - FS-LSTM-4 & 47M & 1.25 \\
        \citet{krause2016multiplicative} - Large mLSTM & 46M & 1.24 \\
        \citet{cmix} - cmix v13 & - & 1.23 \\
        \citet{al2018character} - 12L Transformer & 44M & 1.11 \\
        Ours - 12L Transformer-XL & 41M & \textbf{1.06} \\
        \midrule
        \citet{al2018character} - 64L Transformer & 235M & 1.06 \\
        Ours - 18L Transformer-XL & 88M & 1.03 \\
        Ours - 24L Transformer-XL & 277M & \textbf{0.99} \\
        \bottomrule
    \end{tabular}
    \caption{\small
        Comparison with state-of-the-art results on enwik8.
    }
    \label{table:enwik8}
\end{table}
\egroup

\bgroup
\setlength{\tabcolsep}{3pt}
\begin{table}[t]
    \small
    \centering
    \begin{tabular}{l|cc}
        \toprule
        \bf Model & \bf \#Param & \bf bpc \\
        \midrule
        \citet{cooijmans2016recurrent} - BN-LSTM & - & 1.36 \\
        \citet{chung2016hierarchical} - LN HM-LSTM & 35M & 1.29 \\
        \citet{zilly2016recurrent} - RHN & 45M & 1.27 \\
        \citet{krause2016multiplicative} - Large mLSTM & 45M & 1.27 \\
        \citet{al2018character} - 12L Transformer & 44M & 1.18 \\
        \midrule
        \citet{al2018character} - 64L Transformer & 235M & 1.13 \\
        Ours - 24L Transformer-XL & 277M & \textbf{1.08} \\
        \bottomrule
    \end{tabular}
    \caption{\small
        Comparison with state-of-the-art results on text8.
    }
    \label{table:text8}
\end{table}
\egroup


\bgroup
\setlength{\tabcolsep}{2pt}
\begin{table}[t]
	\small
	\centering
	\begin{adjustwidth}{-4pt}{}
		
	\begin{tabular}{l|ccccc}
		\toprule
		\bf Model & \bf \#Param & \bf PPL \\
		\midrule
		\citet{shazeer2014skip} - Sparse Non-Negative & 33B & 52.9 \\
		\citet{chelba2013one} - RNN-1024 + 9 Gram & 20B & 51.3 \\
		\citet{kuchaiev2017factorization} - G-LSTM-2 & - & 36.0 \\
		\citet{dauphin2016language} - GCNN-14 bottleneck & - & 31.9 \\
		\citet{jozefowicz2016exploring} - LSTM & 1.8B & 30.6 \\
		\citet{jozefowicz2016exploring} - LSTM + CNN Input & 1.04B & 30.0 \\
		\citet{shazeer2017outrageously} - Low-Budget MoE & $\sim$5B & 34.1 \\
		\citet{shazeer2017outrageously} - High-Budget MoE & $\sim$5B & 28.0 \\
		\citet{shazeer2018mesh} - Mesh Tensorflow & 4.9B & 24.0 \\
		\citet{baevski2018adaptive} - Adaptive Input$^\diamond$ & 0.46B & 24.1 \\
		\citet{baevski2018adaptive} - Adaptive Input$^\diamond$ & 1.0B & 23.7 \\
		\midrule
		Ours - Transformer-XL Base & 0.46B & 23.5 \\
		Ours - Transformer-XL Large & 0.8B & \textbf{21.8} \\
		\bottomrule
	\end{tabular}
	\caption{\small
		Comparison with state-of-the-art results on One Billion Word. $^\diamond$ indicates contemporary work.
	}
	\label{table:lm1b}
	\end{adjustwidth}
\end{table}

\begin{table}[t]
    \small
    \centering
    \begin{adjustwidth}{-5pt}{}
    \begin{tabular}{l|cc}
        \toprule
        \bf Model & \bf \#Param & \bf PPL \\
        \midrule
        \citet{inan2016tying} - Tied Variational LSTM & 24M & 73.2 \\
        \citet{zilly2016recurrent} - Variational RHN & 23M & 65.4 \\
        \citet{zoph2016neural} - NAS Cell & 25M & 64.0 \\
        \citet{merity2017regularizing} - AWD-LSTM & 24M & 58.8 \\
        \citet{pham2018efficient} - Efficient NAS & 24M & 58.6 \\
        \citet{liu2018darts} - Differentiable NAS & 23M & 56.1 \\
        \citet{yang2017breaking} - AWD-LSTM-MoS & 22M & 55.97 \\
        \citet{melis2018pushing} - Dropout tuning & 24M & 55.3 \\
        \midrule
        Ours - Transformer-XL & 24M & \textbf{54.52} \\
        \midrule
        \citet{merity2017regularizing} - AWD-LSTM+Finetune$^\dagger$ & 24M & 57.3 \\
        \citet{yang2017breaking} - MoS+Finetune$^\dagger$ & 22M & \textbf{54.44} \\
        \bottomrule
    \end{tabular}
    \caption{\small
        Comparison with state-of-the-art results on Penn Treebank. $\dagger$ indicates using two-step finetuning.
    }
    \label{table:ptb}
	\end{adjustwidth}
\end{table}
\egroup

\begin{table*}[t]
    \small
    \centering
    \begin{tabular}{lcccccc}
        \toprule
        \bf Remark & \bf Recurrence & \bf Encoding & \bf Loss & \bf PPL init & \bf PPL best & \bf Attn Len \\
        \midrule
        Transformer-XL (128M) & \cmark & Ours & Full & \textbf{27.02} & \textbf{26.77} & \textbf{500} \\
        - & \cmark & \citet{shaw2018self} & Full & 27.94 & 27.94 & 256 \\
        - & \cmark & Ours & Half & 28.69 & 28.33 & 460 \\
        - & \xmark & Ours & Full & 29.59 & 29.02 & 260 \\
        - & \xmark & Ours & Half & 30.10 & 30.10 & 120 \\
        \midrule
        - & \xmark & \citet{shaw2018self} & Full & 29.75 & 29.75 & 120 \\
        - & \xmark & \citet{shaw2018self} & Half & 30.50 & 30.50 & 120 \\
        - & \xmark & \citet{vaswani2017attention} & Half & 30.97 & 30.97 & 120 \\
        Transformer (128M)$^\dagger$ & \xmark & \citet{al2018character} & Half & 31.16 & 31.16 & 120 \\
        \midrule
        \multirow{3}{*}{Transformer-XL (151M)} & \multirow{3}{*}{\cmark} & \multirow{3}{*}{Ours} & \multirow{3}{*}{Full} & \multirow{3}{*}{23.43} & \textbf{23.09} & \textbf{640} \\
         &  &  &  &  & 23.16 & 450 \\
          &  &  &  & & 23.35 & 300 \\
        \bottomrule
    \end{tabular}
    \caption{\small
        Ablation study on WikiText-103. For the first two blocks, we use a slightly smaller model (128M parameters). $\dagger$ indicates that the corresponding row is reduced to the same setting as the Transformer network in \cite{al2018character}, except that two auxiliary losses are not implemented in our experiments. ``PPL init'' refers to using the same length as training. ``PPL best'' indicates the perplexity obtained by using the optimal length. ``Attn Len'' is the shortest possible attention length during evaluation to achieve the corresponding result (PPL best).
        Increasing the attention length during evaluation improves performance only when our positional encoding is used.
        The ``Transformer-XL (151M)'' setting uses a standard parameter budget as previous work~\cite{merity2018analysis}, where we observe a similar effect when increasing the attention length during evaluation.
    }
    \label{table:ablation}
\end{table*}

\begin{table}[t]
    \small
    \centering
    \begin{tabular}{lc}
        \toprule
        \bf Method & \bf PPL \\
        \midrule
        Ours & \textbf{25.2} \\
        With \citet{shaw2018self} encodings & 25.7 \\
        Without recurrence & 27.1 \\
        \bottomrule
    \end{tabular}
    \caption{\small
        Ablation study on One Billion Word, a dataset without long-term dependency.
    }
    \label{table:ablation2}
\end{table}

We apply Transformer-XL to a variety of datasets on both word-level and character-level language modeling to have a comparison with state-of-the-art systems, including WikiText-103 \citep{merity2016pointer}, enwik8 \citep{mahoney2011large}, text8~\citep{mahoney2011large}, One Billion Word \citep{chelba2013one}, and Penn Treebank \citep{mikolov2012context}.

WikiText-103 is the largest available word-level language modeling benchmark with long-term dependency. It contains 103M training tokens from 28K articles, with an average length of 3.6K tokens per article, which allows testing the ability of long-term dependency modeling.
We set the attention length to 384 during training and 1600 during evaluation.
We adopted adaptive softmax and input representations \citep{baevski2018adaptive,grave2016efficient}.
As shown in Table \ref{table:103}, Transformer-XL reduces the previous state-of-the-art (SoTA) perplexity from 20.5 to 18.3, which demonstrates the superiority of the Transformer-XL architecture.

The dataset enwik8 contains 100M bytes of unprocessed Wikipedia text.
We compare our architecture with the previous results in Table \ref{table:enwik8}.
Under the model size constraint, the 12-layer Transformer-XL achieves a new SoTA result, outperforming the 12-layer vanilla Transformer from \citet{al2018character} by 0.05, while both Transformer variants have a large margin over conventional RNN-based models.
Notably, our 12-layer architecture achieves the same result as the 64-layer network from \citet{al2018character}, using only 17\% of the parameter budget.
In order to see whether better performances can be obtained by increasing the model size, we train 18-layer and 24-layer Transformer-XLs with increased model sizes.
With the attention length 784 during training and 3,800 during evaluation, we obtained a new SoTA result and our method is the first to break through 1.0 on widely-studied character-level benchmarks.
Different from \citet{al2018character}, Transformer-XL does not need any auxiliary losses, and thus all benefits are credited to a better architecture.

Similar to but different from enwik8, text8 contains 100M processed Wikipedia characters created by lowering case the text and removing any character other than the 26 letters \texttt{a} through \texttt{z}, and space.
Due to the similarity, we simply adapt the best model and the same hyper-parameters on enwik8 to text8 without further tuning.
The comparison with previous methods is summarized in Table \ref{table:text8}.
Again, Transformer-XL achieves the new SoTA result with a clear margin.

One Billion Word does not preserve any long-term dependency because sentences have been shuffled. Consequently, this dataset mainly tests the ability of modeling only short-term dependency.
The comparison between Transformer-XL and the other methods is shown in Table \ref{table:lm1b}.
Although Transformer-XL is mainly designed to better capture longer-term dependency, it dramatically improves the single-model SoTA from 23.7 to 21.8.
Specifically, Transformer-XL significantly outperforms a contemporary method using vanilla Transformers~\cite{baevski2018adaptive}, suggesting the advantage of Transformer-XL is generalizable to modeling short sequences.


We also report the results on word-level Penn Treebank in Table \ref{table:ptb}. Similar to AWD-LSTM \citep{merity2017regularizing}, we apply variational dropout and weight average to Transformer-XL. With proper regularization, Transformer-XL achieves a new SoTA result among models without two-step finetuning. Penn Treebank has only 1M training tokens, which implies that Transformer-XL also generalizes well even on small datasets.

\subsection{Ablation Study}


We conduct two sets of ablation studies to examine the effects of two proposed techniques used in Transformer-XL: the recurrence mechanism and the new positional encoding scheme.

The first study is performed on WikiText-103, which requires modeling long-term dependency.
The results are reported in Table \ref{table:ablation}.
Among the compared encoding schemes, \citet{shaw2018self} is relative, while \citet{vaswani2017attention} and \citet{al2018character} are absolute. ``Full'' and ``half'' losses refer to applying a cross entropy loss to all or the recent half positions in the segment.
We found that absolute encodings only work well with half losses because half losses exclude positions with very short attention lengths during training for better generalization.
Table \ref{table:ablation} shows that both the recurrence mechanism and our encoding scheme are necessary to achieve the best performance, as well as generalizing to longer attention sequences during evaluation time. Although the backpropagation length during training is only 128, with the two techniques the attention length can be increased to 640 at test time. In the standard setting with 151M parameters, the perplexity decreases as the attention length increases.

Since the recurrence mechanism costs additional memory, we also compare Transformer-XL with baselines under the same GPU memory constraints. As shown in Table \ref{table:memory} in Appendix \ref{sec:memory}, despite using a shorter backpropagation length, Transformer-XL remains superior to the baselines.

The second study targets at isolating the effects of resolving the context fragmentation problem from the benefit of capturing longer context length.
In order to achieve this goal, we deliberately choose a dataset that does not require long-term dependency, so that any improvement from establishing the recurrence can be attributed to solving the context fragmentation.
Specifically, we perform this controlled experiment on the One Billion Word dataset, which can only benefit from removing the context fragmentation. We train a 20-layer Transformer-XL with $\sim$0.3B parameters for 400K steps.
As shown in Table \ref{table:ablation2}, using segment-level recurrence substantially improves performance even when long-term dependency is not needed, which is consistent with our previous discussion that the recurrence mechanism resolves the context fragmentation problem. Moreover, our relative positional encodings is also superior to \citet{shaw2018self} on short sequences.

\subsection{Relative Effective Context Length}
\bgroup
\setlength{\tabcolsep}{2pt}
\begin{table}[t]
	\small
	\centering
	\begin{tabular}{lccc}
		\toprule
		\bf Model & $r=0.1$ & $r=0.5$ & $r=1.0$ \\
		\midrule
		Transformer-XL 151M & \textbf{900} & \textbf{800} & \textbf{700} \\
		QRNN & 500 & 400 & 300 \\
		LSTM & 400 & 300 & 200 \\
		\midrule
		Transformer-XL 128M & \textbf{700} & \textbf{600} & \textbf{500} \\
		- use \citet{shaw2018self} encoding & 400 & 400 & 300 \\
		- remove recurrence & 300 & 300 & 300 \\
		Transformer & 128 & 128 & 128 \\
		\bottomrule
	\end{tabular}
	\caption{\small
		Relative effective context length (RECL) comparison. See text for the definition of RECL and $r$. The first three models and the last four models are
		compared as two \textit{model groups} when we calculate RECL (RECL is computed on a model group rather than a single model). Each group has the same parameter budget.
	}
	\label{table:recl}
\end{table}
\egroup

\citet{khandelwal2018sharp} proposed a method to evaluate the \textit{Effective Context Length} (ECL) of a sequence model.
ECL is the longest length to which increasing the context span would lead to a gain more than a threshold.
However, ECL ignores the fact that it is harder to get improvement when a model already achieves a lower perplexity using only a shorter context, and thus it is not suitable for fair comparison among multiple models. We instead propose a new metric called \textit{Relative Effective Context Length} (RECL). RECL is defined on a model group instead of a single model, and the gain of a long context is measure by the relative improvement over the \textit{best} short context model. As such, the model group shares the same baseline to enable fair comparison. RECL also has a parameter $r$, which means constraining the comparison on top-$r$ hard examples. See Appedix \ref{sec:recl} for more details about RECL. As shown in Table \ref{table:recl}, Transformer-XL manages to model dependency of 900 words long on average with $r = 0.1$. The RECL of Transformer-XL is 80\% and 450\% longer than recurrent networks and Transformer respectively. Both the recurrence mechanism and our positional encodings contribute to a longer RECL. This further substantiates our argument that Transformer-XL is able to model longer-term dependency.

\subsection{Generated Text}

Trained only on WikiText-103 which is medium-sized, Transformer-XL is already able to generate relatively coherent articles with thousands of tokens without manual cherry picking, despite minor flaws. Please refer to Appendix \ref{sec:gen} for samples.

\subsection{Evaluation Speed} \label{sec:speed}
Finally, we compare the evaluation speed of our model with the vanilla Transformer model~\cite{al2018character}. As shown in Table \ref{table:speed}, due to the state reuse scheme, Transformer-XL achieves an up to 1,874 times speedup during evaluation.

\begin{table}[t]
	\small
	\centering
	\begin{tabular}{cc}
		\toprule
		\bf Attn Len & \bf How much \citet{al2018character} is slower \\
		\midrule
		3,800 & 1,874x \\
		2,800 & 1,409x \\
		1,800 & 773x \\
		800 & 363x \\
		\bottomrule
	\end{tabular}
	\caption{\small
		Slowdown in terms of running time during evaluation. Evaluation is based on per-token time on one GPU.
	}
	\label{table:speed}
\end{table}

\section{Conclusions}

Transformer-XL obtains strong perplexity results, models longer-term dependency than RNNs and Transformer, achieves substantial speedup during evaluation, and is able to generate coherent text articles. We envision interesting applications of Transformer-XL in the fields of text generation, unsupervised feature learning, image and speech modeling.

\subsubsection*{Acknowledgments}
ZD and YY were supported in part by National Science Foundation (NSF) under the grant IIS-1546329 and by the DOE-Office of Science under the grant ASCR \#KJ040201.
ZY and RS were supported in part by the Office of Naval Research grant N000141812861, the NSF grant IIS1763562, the Nvidia fellowship, and the Siebel scholarship.

\FloatBarrier

\bibliography{acl2019}

\begin{thebibliography}{65}
\expandafter\ifx\csname natexlab\endcsname\relax\def\natexlab#1{#1}\fi

\bibitem[{Al-Rfou et~al.(2018)Al-Rfou, Choe, Constant, Guo, and
  Jones}]{al2018character}
Rami Al-Rfou, Dokook Choe, Noah Constant, Mandy Guo, and Llion Jones. 2018.
\newblock Character-level language modeling with deeper self-attention.
\newblock \emph{arXiv preprint arXiv:1808.04444}.

\bibitem[{Baevski and Auli(2018)}]{baevski2018adaptive}
Alexei Baevski and Michael Auli. 2018.
\newblock Adaptive input representations for neural language modeling.
\newblock \emph{arXiv preprint arXiv:1809.10853}.

\bibitem[{Bahdanau et~al.(2014)Bahdanau, Cho, and Bengio}]{bahdanau2014neural}
Dzmitry Bahdanau, Kyunghyun Cho, and Yoshua Bengio. 2014.
\newblock Neural machine translation by jointly learning to align and
  translate.
\newblock \emph{arXiv preprint arXiv:1409.0473}.

\bibitem[{Bai et~al.(2018)Bai, Kolter, and Koltun}]{bai2018empirical}
Shaojie Bai, J~Zico Kolter, and Vladlen Koltun. 2018.
\newblock An empirical evaluation of generic convolutional and recurrent
  networks for sequence modeling.
\newblock \emph{arXiv preprint arXiv:1803.01271}.

\bibitem[{Bengio et~al.(2003)Bengio, Ducharme, Vincent, and
  Jauvin}]{bengio2003neural}
Yoshua Bengio, R{\'e}jean Ducharme, Pascal Vincent, and Christian Jauvin. 2003.
\newblock A neural probabilistic language model.
\newblock \emph{Journal of machine learning research}, 3(Feb):1137--1155.

\bibitem[{Chelba et~al.(2013)Chelba, Mikolov, Schuster, Ge, Brants, Koehn, and
  Robinson}]{chelba2013one}
Ciprian Chelba, Tomas Mikolov, Mike Schuster, Qi~Ge, Thorsten Brants, Phillipp
  Koehn, and Tony Robinson. 2013.
\newblock One billion word benchmark for measuring progress in statistical
  language modeling.
\newblock \emph{arXiv preprint arXiv:1312.3005}.

\bibitem[{Chung et~al.(2016)Chung, Ahn, and Bengio}]{chung2016hierarchical}
Junyoung Chung, Sungjin Ahn, and Yoshua Bengio. 2016.
\newblock Hierarchical multiscale recurrent neural networks.
\newblock \emph{arXiv preprint arXiv:1609.01704}.

\bibitem[{Cooijmans et~al.(2016)Cooijmans, Ballas, Laurent, G{\"u}l{\c{c}}ehre,
  and Courville}]{cooijmans2016recurrent}
Tim Cooijmans, Nicolas Ballas, C{\'e}sar Laurent, {\c{C}}a{\u{g}}lar
  G{\"u}l{\c{c}}ehre, and Aaron Courville. 2016.
\newblock Recurrent batch normalization.
\newblock \emph{arXiv preprint arXiv:1603.09025}.

\bibitem[{Dai and Le(2015)}]{dai2015semi}
Andrew~M Dai and Quoc~V Le. 2015.
\newblock Semi-supervised sequence learning.
\newblock In \emph{Advances in neural information processing systems}, pages
  3079--3087.

\bibitem[{Dauphin et~al.(2016)Dauphin, Fan, Auli, and
  Grangier}]{dauphin2016language}
Yann~N Dauphin, Angela Fan, Michael Auli, and David Grangier. 2016.
\newblock Language modeling with gated convolutional networks.
\newblock \emph{arXiv preprint arXiv:1612.08083}.

\bibitem[{Devlin et~al.(2018)Devlin, Chang, Lee, and
  Toutanova}]{devlin2018bert}
Jacob Devlin, Ming-Wei Chang, Kenton Lee, and Kristina Toutanova. 2018.
\newblock Bert: Pre-training of deep bidirectional transformers for language
  understanding.
\newblock \emph{arXiv preprint arXiv:1810.04805}.

\bibitem[{Dieng et~al.(2016)Dieng, Wang, Gao, and Paisley}]{dieng2016topicrnn}
Adji~B Dieng, Chong Wang, Jianfeng Gao, and John Paisley. 2016.
\newblock Topicrnn: A recurrent neural network with long-range semantic
  dependency.
\newblock \emph{arXiv preprint arXiv:1611.01702}.

\bibitem[{Gal and Ghahramani(2016)}]{gal2016theoretically}
Yarin Gal and Zoubin Ghahramani. 2016.
\newblock A theoretically grounded application of dropout in recurrent neural
  networks.
\newblock In \emph{Advances in neural information processing systems}, pages
  1019--1027.

\bibitem[{Grave et~al.(2016{\natexlab{a}})Grave, Joulin, Ciss{\'e}, Grangier,
  and J{\'e}gou}]{grave2016efficient}
Edouard Grave, Armand Joulin, Moustapha Ciss{\'e}, David Grangier, and
  Herv{\'e} J{\'e}gou. 2016{\natexlab{a}}.
\newblock Efficient softmax approximation for gpus.
\newblock \emph{arXiv preprint arXiv:1609.04309}.

\bibitem[{Grave et~al.(2016{\natexlab{b}})Grave, Joulin, and
  Usunier}]{grave2016improving}
Edouard Grave, Armand Joulin, and Nicolas Usunier. 2016{\natexlab{b}}.
\newblock Improving neural language models with a continuous cache.
\newblock \emph{arXiv preprint arXiv:1612.04426}.

\bibitem[{Graves(2013)}]{graves2013generating}
Alex Graves. 2013.
\newblock Generating sequences with recurrent neural networks.
\newblock \emph{arXiv preprint arXiv:1308.0850}.

\bibitem[{Graves et~al.(2014)Graves, Wayne, and Danihelka}]{graves2014neural}
Alex Graves, Greg Wayne, and Ivo Danihelka. 2014.
\newblock Neural turing machines.
\newblock \emph{arXiv preprint arXiv:1410.5401}.

\bibitem[{Ha et~al.(2016)Ha, Dai, and Le}]{ha2016hypernetworks}
David Ha, Andrew Dai, and Quoc~V Le. 2016.
\newblock Hypernetworks.
\newblock \emph{arXiv preprint arXiv:1609.09106}.

\bibitem[{Hochreiter et~al.(2001)Hochreiter, Bengio, Frasconi, Schmidhuber
  et~al.}]{hochreiter2001gradient}
Sepp Hochreiter, Yoshua Bengio, Paolo Frasconi, J{\"u}rgen Schmidhuber, et~al.
  2001.
\newblock Gradient flow in recurrent nets: the difficulty of learning long-term
  dependencies.

\bibitem[{Hochreiter and Schmidhuber(1997)}]{hochreiter1997long}
Sepp Hochreiter and J{\"u}rgen Schmidhuber. 1997.
\newblock Long short-term memory.
\newblock \emph{Neural computation}, 9(8):1735--1780.

\bibitem[{Huang et~al.(2018)Huang, Vaswani, Uszkoreit, Shazeer, Hawthorne, Dai,
  Hoffman, and Eck}]{huang2018improved}
Cheng-Zhi~Anna Huang, Ashish Vaswani, Jakob Uszkoreit, Noam Shazeer, Curtis
  Hawthorne, Andrew~M Dai, Matthew~D Hoffman, and Douglas Eck. 2018.
\newblock An improved relative self-attention mechanism for transformer with
  application to music generation.
\newblock \emph{arXiv preprint arXiv:1809.04281}.

\bibitem[{Inan et~al.(2016)Inan, Khosravi, and Socher}]{inan2016tying}
Hakan Inan, Khashayar Khosravi, and Richard Socher. 2016.
\newblock Tying word vectors and word classifiers: A loss framework for
  language modeling.
\newblock \emph{arXiv preprint arXiv:1611.01462}.

\bibitem[{Ji et~al.(2015)Ji, Cohn, Kong, Dyer, and Eisenstein}]{ji2015document}
Yangfeng Ji, Trevor Cohn, Lingpeng Kong, Chris Dyer, and Jacob Eisenstein.
  2015.
\newblock Document context language models.
\newblock \emph{arXiv preprint arXiv:1511.03962}.

\bibitem[{Jozefowicz et~al.(2016)Jozefowicz, Vinyals, Schuster, Shazeer, and
  Wu}]{jozefowicz2016exploring}
Rafal Jozefowicz, Oriol Vinyals, Mike Schuster, Noam Shazeer, and Yonghui Wu.
  2016.
\newblock Exploring the limits of language modeling.
\newblock \emph{arXiv preprint arXiv:1602.02410}.

\bibitem[{Kalchbrenner et~al.(2016)Kalchbrenner, Espeholt, Simonyan, Oord,
  Graves, and Kavukcuoglu}]{kalchbrenner2016neural}
Nal Kalchbrenner, Lasse Espeholt, Karen Simonyan, Aaron van~den Oord, Alex
  Graves, and Koray Kavukcuoglu. 2016.
\newblock Neural machine translation in linear time.
\newblock \emph{arXiv preprint arXiv:1610.10099}.

\bibitem[{Kanai et~al.(2018)Kanai, Fujiwara, Yamanaka, and
  Adachi}]{kanai2018sigsoftmax}
Sekitoshi Kanai, Yasuhiro Fujiwara, Yuki Yamanaka, and Shuichi Adachi. 2018.
\newblock Sigsoftmax: Reanalysis of the softmax bottleneck.
\newblock \emph{arXiv preprint arXiv:1805.10829}.

\bibitem[{Ke et~al.(2018)Ke, GOYAL, Bilaniuk, Binas, Mozer, Pal, and
  Bengio}]{ke2018sparse}
Nan~Rosemary Ke, Anirudh Goyal ALIAS~PARTH GOYAL, Olexa Bilaniuk, Jonathan
  Binas, Michael~C Mozer, Chris Pal, and Yoshua Bengio. 2018.
\newblock Sparse attentive backtracking: Temporal credit assignment through
  reminding.
\newblock In \emph{Advances in Neural Information Processing Systems}, pages
  7650--7661.

\bibitem[{Khandelwal et~al.(2018)Khandelwal, He, Qi, and
  Jurafsky}]{khandelwal2018sharp}
Urvashi Khandelwal, He~He, Peng Qi, and Dan Jurafsky. 2018.
\newblock Sharp nearby, fuzzy far away: How neural language models use context.
\newblock \emph{arXiv preprint arXiv:1805.04623}.

\bibitem[{Knol(2017)}]{cmix}
Bryon Knol. 2017.
\newblock cmix v13.
\newblock \url{http://www.byronknoll.com/cmix.html}.

\bibitem[{Koutnik et~al.(2014)Koutnik, Greff, Gomez, and
  Schmidhuber}]{koutnik2014clockwork}
Jan Koutnik, Klaus Greff, Faustino Gomez, and Juergen Schmidhuber. 2014.
\newblock A clockwork rnn.
\newblock \emph{arXiv preprint arXiv:1402.3511}.

\bibitem[{Krause et~al.(2016)Krause, Lu, Murray, and
  Renals}]{krause2016multiplicative}
Ben Krause, Liang Lu, Iain Murray, and Steve Renals. 2016.
\newblock Multiplicative lstm for sequence modelling.
\newblock \emph{arXiv preprint arXiv:1609.07959}.

\bibitem[{Kuchaiev and Ginsburg(2017)}]{kuchaiev2017factorization}
Oleksii Kuchaiev and Boris Ginsburg. 2017.
\newblock Factorization tricks for lstm networks.
\newblock \emph{arXiv preprint arXiv:1703.10722}.

\bibitem[{Le et~al.(2015)Le, Jaitly, and Hinton}]{le2015simple}
Quoc~V Le, Navdeep Jaitly, and Geoffrey~E Hinton. 2015.
\newblock A simple way to initialize recurrent networks of rectified linear
  units.
\newblock \emph{arXiv preprint arXiv:1504.00941}.

\bibitem[{Li et~al.(2018)Li, Li, Cook, Zhu, and Gao}]{li2018independently}
Shuai Li, Wanqing Li, Chris Cook, Ce~Zhu, and Yanbo Gao. 2018.
\newblock Independently recurrent neural network (indrnn): Building a longer
  and deeper rnn.
\newblock In \emph{Proceedings of the IEEE Conference on Computer Vision and
  Pattern Recognition}, pages 5457--5466.

\bibitem[{Liu et~al.(2018)Liu, Simonyan, and Yang}]{liu2018darts}
Hanxiao Liu, Karen Simonyan, and Yiming Yang. 2018.
\newblock Darts: Differentiable architecture search.
\newblock \emph{arXiv preprint arXiv:1806.09055}.

\bibitem[{LLC(2009)}]{mahoney2011large}
MultiMedia LLC. 2009.
\newblock Large text compression benchmark.

\bibitem[{Melis et~al.(2018)Melis, Blundell, Ko{\v{c}}isk{\`y}, Hermann, Dyer,
  and Blunsom}]{melis2018pushing}
G{\'a}bor Melis, Charles Blundell, Tom{\'a}{\v{s}} Ko{\v{c}}isk{\`y},
  Karl~Moritz Hermann, Chris Dyer, and Phil Blunsom. 2018.
\newblock Pushing the bounds of dropout.
\newblock \emph{arXiv preprint arXiv:1805.09208}.

\bibitem[{Merity et~al.(2017)Merity, Keskar, and
  Socher}]{merity2017regularizing}
Stephen Merity, Nitish~Shirish Keskar, and Richard Socher. 2017.
\newblock Regularizing and optimizing lstm language models.
\newblock \emph{arXiv preprint arXiv:1708.02182}.

\bibitem[{Merity et~al.(2018)Merity, Keskar, and Socher}]{merity2018analysis}
Stephen Merity, Nitish~Shirish Keskar, and Richard Socher. 2018.
\newblock An analysis of neural language modeling at multiple scales.
\newblock \emph{arXiv preprint arXiv:1803.08240}.

\bibitem[{Merity et~al.(2016)Merity, Xiong, Bradbury, and
  Socher}]{merity2016pointer}
Stephen Merity, Caiming Xiong, James Bradbury, and Richard Socher. 2016.
\newblock Pointer sentinel mixture models.
\newblock \emph{arXiv preprint arXiv:1609.07843}.

\bibitem[{Mikolov et~al.(2014)Mikolov, Joulin, Chopra, Mathieu, and
  Ranzato}]{mikolov2014learning}
Tomas Mikolov, Armand Joulin, Sumit Chopra, Michael Mathieu, and Marc'Aurelio
  Ranzato. 2014.
\newblock Learning longer memory in recurrent neural networks.
\newblock \emph{arXiv preprint arXiv:1412.7753}.

\bibitem[{Mikolov et~al.(2010)Mikolov, Karafi{\'a}t, Burget,
  {\v{C}}ernock{\`y}, and Khudanpur}]{mikolov2010recurrent}
Tom{\'a}{\v{s}} Mikolov, Martin Karafi{\'a}t, Luk{\'a}{\v{s}} Burget, Jan
  {\v{C}}ernock{\`y}, and Sanjeev Khudanpur. 2010.
\newblock Recurrent neural network based language model.
\newblock In \emph{Eleventh Annual Conference of the International Speech
  Communication Association}.

\bibitem[{Mikolov and Zweig(2012)}]{mikolov2012context}
Tomas Mikolov and Geoffrey Zweig. 2012.
\newblock Context dependent recurrent neural network language model.
\newblock \emph{SLT}, 12(234-239):8.

\bibitem[{Morin and Bengio(2005)}]{morin2005hierarchical}
Frederic Morin and Yoshua Bengio. 2005.
\newblock Hierarchical probabilistic neural network language model.
\newblock In \emph{Aistats}, volume~5, pages 246--252. Citeseer.

\bibitem[{Mujika et~al.(2017)Mujika, Meier, and Steger}]{mujika2017fast}
Asier Mujika, Florian Meier, and Angelika Steger. 2017.
\newblock Fast-slow recurrent neural networks.
\newblock In \emph{Advances in Neural Information Processing Systems}, pages
  5915--5924.

\bibitem[{Pascanu et~al.(2012)Pascanu, Mikolov, and
  Bengio}]{pascanu2012understanding}
Razvan Pascanu, Tomas Mikolov, and Yoshua Bengio. 2012.
\newblock Understanding the exploding gradient problem.
\newblock \emph{CoRR, abs/1211.5063}.

\bibitem[{Peters et~al.(2018)Peters, Neumann, Iyyer, Gardner, Clark, Lee, and
  Zettlemoyer}]{peters2018deep}
Matthew~E Peters, Mark Neumann, Mohit Iyyer, Matt Gardner, Christopher Clark,
  Kenton Lee, and Luke Zettlemoyer. 2018.
\newblock Deep contextualized word representations.
\newblock \emph{arXiv preprint arXiv:1802.05365}.

\bibitem[{Pham et~al.(2018)Pham, Guan, Zoph, Le, and Dean}]{pham2018efficient}
Hieu Pham, Melody~Y Guan, Barret Zoph, Quoc~V Le, and Jeff Dean. 2018.
\newblock Efficient neural architecture search via parameter sharing.
\newblock \emph{arXiv preprint arXiv:1802.03268}.

\bibitem[{Press and Wolf(2016)}]{press2016using}
Ofir Press and Lior Wolf. 2016.
\newblock Using the output embedding to improve language models.
\newblock \emph{arXiv preprint arXiv:1608.05859}.

\bibitem[{Radford et~al.(2018)Radford, Narasimhan, Salimans, and
  Sutskever}]{radford2018improving}
Alec Radford, Karthik Narasimhan, Tim Salimans, and Ilya Sutskever. 2018.
\newblock Improving language understanding by generative pre-training.
\newblock \emph{URL https://s3-us-west-2. amazonaws.
  com/openai-assets/research-covers/languageunsupervised/language understanding
  paper. pdf}.

\bibitem[{Rae et~al.(2018)Rae, Dyer, Dayan, and Lillicrap}]{rae2018fast}
Jack~W Rae, Chris Dyer, Peter Dayan, and Timothy~P Lillicrap. 2018.
\newblock Fast parametric learning with activation memorization.
\newblock \emph{arXiv preprint arXiv:1803.10049}.

\bibitem[{Shaw et~al.(2018)Shaw, Uszkoreit, and Vaswani}]{shaw2018self}
Peter Shaw, Jakob Uszkoreit, and Ashish Vaswani. 2018.
\newblock Self-attention with relative position representations.
\newblock \emph{arXiv preprint arXiv:1803.02155}.

\bibitem[{Shazeer et~al.(2018)Shazeer, Cheng, Parmar, Tran, Vaswani,
  Koanantakool, Hawkins, Lee, Hong, Young et~al.}]{shazeer2018mesh}
Noam Shazeer, Youlong Cheng, Niki Parmar, Dustin Tran, Ashish Vaswani, Penporn
  Koanantakool, Peter Hawkins, HyoukJoong Lee, Mingsheng Hong, Cliff Young,
  et~al. 2018.
\newblock Mesh-tensorflow: Deep learning for supercomputers.
\newblock In \emph{Advances in Neural Information Processing Systems}, pages
  10434--10443.

\bibitem[{Shazeer et~al.(2017)Shazeer, Mirhoseini, Maziarz, Davis, Le, Hinton,
  and Dean}]{shazeer2017outrageously}
Noam Shazeer, Azalia Mirhoseini, Krzysztof Maziarz, Andy Davis, Quoc Le,
  Geoffrey Hinton, and Jeff Dean. 2017.
\newblock Outrageously large neural networks: The sparsely-gated
  mixture-of-experts layer.
\newblock \emph{arXiv preprint arXiv:1701.06538}.

\bibitem[{Shazeer et~al.(2014)Shazeer, Pelemans, and Chelba}]{shazeer2014skip}
Noam Shazeer, Joris Pelemans, and Ciprian Chelba. 2014.
\newblock Skip-gram language modeling using sparse non-negative matrix
  probability estimation.
\newblock \emph{arXiv preprint arXiv:1412.1454}.

\bibitem[{Trinh et~al.(2018)Trinh, Dai, Luong, and Le}]{trinh2018learning}
Trieu~H Trinh, Andrew~M Dai, Thang Luong, and Quoc~V Le. 2018.
\newblock Learning longer-term dependencies in rnns with auxiliary losses.
\newblock \emph{arXiv preprint arXiv:1803.00144}.

\bibitem[{Vaswani et~al.(2017)Vaswani, Shazeer, Parmar, Uszkoreit, Jones,
  Gomez, Kaiser, and Polosukhin}]{vaswani2017attention}
Ashish Vaswani, Noam Shazeer, Niki Parmar, Jakob Uszkoreit, Llion Jones,
  Aidan~N Gomez, {\L}ukasz Kaiser, and Illia Polosukhin. 2017.
\newblock Attention is all you need.
\newblock In \emph{Advances in Neural Information Processing Systems}, pages
  5998--6008.

\bibitem[{Wang and Cho(2015)}]{wang2015larger}
Tian Wang and Kyunghyun Cho. 2015.
\newblock Larger-context language modelling.
\newblock \emph{arXiv preprint arXiv:1511.03729}.

\bibitem[{Wang et~al.(2017)Wang, Gan, Wang, Shen, Huang, Ping, Satheesh, and
  Carin}]{wang2017topic}
Wenlin Wang, Zhe Gan, Wenqi Wang, Dinghan Shen, Jiaji Huang, Wei Ping, Sanjeev
  Satheesh, and Lawrence Carin. 2017.
\newblock Topic compositional neural language model.
\newblock \emph{arXiv preprint arXiv:1712.09783}.

\bibitem[{Weston et~al.(2014)Weston, Chopra, and Bordes}]{weston2014memory}
Jason Weston, Sumit Chopra, and Antoine Bordes. 2014.
\newblock Memory networks.
\newblock \emph{arXiv preprint arXiv:1410.3916}.

\bibitem[{Wu et~al.(2016)Wu, Zhang, Zhang, Bengio, and
  Salakhutdinov}]{wu2016multiplicative}
Yuhuai Wu, Saizheng Zhang, Ying Zhang, Yoshua Bengio, and Ruslan~R
  Salakhutdinov. 2016.
\newblock On multiplicative integration with recurrent neural networks.
\newblock In \emph{Advances in neural information processing systems}, pages
  2856--2864.

\bibitem[{Yang et~al.(2017)Yang, Dai, Salakhutdinov, and
  Cohen}]{yang2017breaking}
Zhilin Yang, Zihang Dai, Ruslan Salakhutdinov, and William~W Cohen. 2017.
\newblock Breaking the softmax bottleneck: A high-rank rnn language model.
\newblock \emph{arXiv preprint arXiv:1711.03953}.

\bibitem[{Zaremba et~al.(2014)Zaremba, Sutskever, and
  Vinyals}]{zaremba2014recurrent}
Wojciech Zaremba, Ilya Sutskever, and Oriol Vinyals. 2014.
\newblock Recurrent neural network regularization.
\newblock \emph{arXiv preprint arXiv:1409.2329}.

\bibitem[{Zilly et~al.(2016)Zilly, Srivastava, Koutn{\'\i}k, and
  Schmidhuber}]{zilly2016recurrent}
Julian~Georg Zilly, Rupesh~Kumar Srivastava, Jan Koutn{\'\i}k, and J{\"u}rgen
  Schmidhuber. 2016.
\newblock Recurrent highway networks.
\newblock \emph{arXiv preprint arXiv:1607.03474}.

\bibitem[{Zoph and Le(2016)}]{zoph2016neural}
Barret Zoph and Quoc~V Le. 2016.
\newblock Neural architecture search with reinforcement learning.
\newblock \emph{arXiv preprint arXiv:1611.01578}.

\end{thebibliography}
\bibliographystyle{acl_natbib}

\FloatBarrier

\appendix
\onecolumn
\section{Ablation Study with Memory Constraints} \label{sec:memory}

\begin{table}[!h]
    \small
    \centering
    \begin{tabular}{ccccccc}
        \toprule
        \bf Backprop Len & \bf Recurrence & \bf Encoding & \bf Loss & \bf pplx best & \bf pplx init & \bf Attn Len \\
        \midrule
        128 & \cmark & Ours & Full & \textbf{26.77} & \textbf{27.02} & \textbf{500} \\
        128 & \cmark & Ours & Partial & 28.33 & 28.69 & 460 \\
        \midrule
        176 & \xmark & Ours & Full & 27.98 & 28.43 & 400 \\
        172 & \xmark & Ours & Partial & 28.83 & 28.83 & 120 \\
        \bottomrule
    \end{tabular}
    \caption{\small
        Ablation study on WikiText-103 with the same GPU memory constraints.
    }
    \label{table:memory}
\end{table}

Table \ref{table:memory} compares Transformer-XL with baseline under the same memory budget. Transformer-XL still outperforms the baseline even with a shorter backprop length.

\section{Efficient Computation of the Attention with Relative Positional Embedding}
\label{sec:A-efficient-attention}
As we discussed in section \ref{sec:rel-pos-embed}, the naive way of computing the $\rmW_{k,R} \rmR_{i-j}$ for all pairs $(i, j)$ is subject to a quadratic cost.
Here, we present a simple method with only a linear cost.
Firstly, notice that the relative distance $i - j$ can only be integer from 0 to $M+L-1$, where $M$ and $L$ are the memory length and segment length respectively.
Hence, the rows of the matrix
\[\small
	\rmQ
	\coloneqq \begin{bmatrix} \rmR_{M+L-1}^\top \\ \rmR_{M+L-2}^\top \\ \vdots \\ \rmR_{1}^\top \\ \rmR_{0}^\top \end{bmatrix} {\rmW_{k,R}}^\top
	= \begin{bmatrix}
		\left[ \rmW_{k,R} \rmR_{M+L-1} \right]^\top \\
		\left[ \rmW_{k,R} \rmR_{M+L-2} \right]^\top \\ \vdots \\
		\left[ \rmW_{k,R} \rmR_{1} \right]^\top \\
		\left[ \rmW_{k,R} \rmR_{0} \right]^\top \end{bmatrix}
	\in \R^{(M+L) \times d}
\]
consist of all possible vector outputs of $\rmW_{k,R} \rmR_{i-j}$ for any $(i, j)$.
Note that we have defined $\rmQ$ in a reversed order, i.e., $\rmQ_k = \rmW_{k,R} \rmR_{M+L-1-k}$, to make further discussion easier.

Next, we collect the term $(b)$ for all possible $i, j$ into the following $L \times (M+L)$ matrix,
\begin{align*}
\rmB &=
	\begin{bmatrix}
	q_0^\top \rmW_{k,R} \rmR_{M}   & \cdots & q_0^\top \rmW_{k,R} \rmR_{0} & 0 & \cdots & 0 \\
	q_1^\top \rmW_{k,R} \rmR_{M+1} & \cdots & q_1^\top \rmW_{k,R} \rmR_{1} & q_1^\top \rmW_{k,R} \rmR_{0} & \cdots & 0 \\
	\vdots & \vdots & \vdots & \vdots & \ddots & \vdots \\
	q_{L-1}^\top \rmW_{k,R} \rmR_{M+L-1} & \cdots & q_{L-1}^\top \rmW_{k,R} \rmR_{M+L-1} & q_{L-1}^\top \rmW_{k,R} \rmR_{L-1} & \cdots & q_{L-1}^\top \rmW_{k,R} \rmR_{0} \\
	\end{bmatrix} \\
	&=
	\begin{bmatrix}
	q_0^\top \rmQ_{L-1}   & \cdots & q_0^\top \rmQ_{M+L-1} & 0                       & \cdots & 0 \\
	q_1^\top \rmQ_{L-2}   & \cdots & q_1^\top \rmQ_{M+L-2} & q_1^\top \rmQ_{M+L-1}   & \cdots & 0 \\
	\vdots                & \vdots & \ddots                & \vdots                  & \ddots & \vdots \\
	q_{L-1}^\top \rmQ_{0} & \cdots & q_{L-1}^\top \rmQ_{M} & q_{L-1}^\top \rmQ_{M+1} & \cdots & q_{L-1}^\top \rmQ_{M+L-1} \\
	\end{bmatrix}
\end{align*}
Then, we further define
\[
	\widetilde{\rmB} = \rvq \rmQ^\top =
	\begin{bmatrix}
	q_{0}^\top \rmQ_{0}   & \cdots & q_{0}^\top \rmQ_{M}   & q_{0}^\top \rmQ_{M+1}    & \cdots & q_{0}^\top \rmQ_{M+L-1}   \\
	q_{1}^\top \rmQ_{0}   & \cdots & q_{1}^\top \rmQ_{M}   & q_{1}^\top \rmQ_{M+1}    & \cdots & q_{1}^\top \rmQ_{M+L-1}   \\
	\vdots                & \vdots & \ddots                & \vdots                   & \ddots & \vdots                    \\
	q_{L-1}^\top \rmQ_{0} & \cdots & q_{L-1}^\top \rmQ_{M} & q_{L-1}^\top \rmQ_{M+1}  & \cdots & q_{L-1}^\top \rmQ_{M+L-1} \\
	\end{bmatrix}.
\]
Now, it is easy to see an immediate relationship between $\rmB$ and $\widetilde{\rmB}$, where the $i$-th row of $\rmB$ is simply a left-shifted version of $i$-th row of $\widetilde{\rmB}$.
Hence, the computation of $\rmB$ only requires a matrix multiplication $\rvq \rmQ^\top$ to compute $\widetilde{\rmB}$ and then a set of left-shifts.

Similarly, we can collect all term $(d)$ for all possible $i, j$ into another $L \times (M+L)$ matrix $\rmD$,
\begin{align*}
	\rmD &=
	\begin{bmatrix}
	v^\top \rmQ_{L-1} & \cdots & v^\top \rmQ_{M+L-1} & 0 & \cdots & 0 \\
	v^\top \rmQ_{L-2} & \cdots & v^\top \rmQ_{M+L-2} & v^\top \rmQ_{M+L-1} & \cdots & 0 \\
	\vdots            & \vdots & \ddots              & \vdots              & \ddots & \vdots     \\
	v^\top \rmQ_{0}   & \cdots & v^\top \rmQ_{M}     & v^\top \rmQ_{M+1}   & \cdots & v^\top \rmQ_{M+L-1} \\
	\end{bmatrix}.
\end{align*}
Then, we can follow the same procedure to define
\[
\widetilde{\rvd} = \left[\rmQ v \right]^\top
	=
	\begin{bmatrix}
	v^\top \rmQ_{0}   & \cdots & v^\top \rmQ_{M}     & v^\top \rmQ_{M+1}   & \cdots & v^\top \rmQ_{M+L-1}
	\end{bmatrix}.
\]
Again, each row of $\rmD$ is simply a left-shift version of $\widetilde{\rvd}$.
Hence, the main computation cost comes from the matrix-vector multiplication $\widetilde{\rvd} = \left[\rmQ v \right]^\top$, which is not expensive any more.

\section{Details About RECL} \label{sec:recl}
\begin{figure}[!h]
	\begin{subfigure}[b]{0.5\textwidth}
		\includegraphics[width=\textwidth]{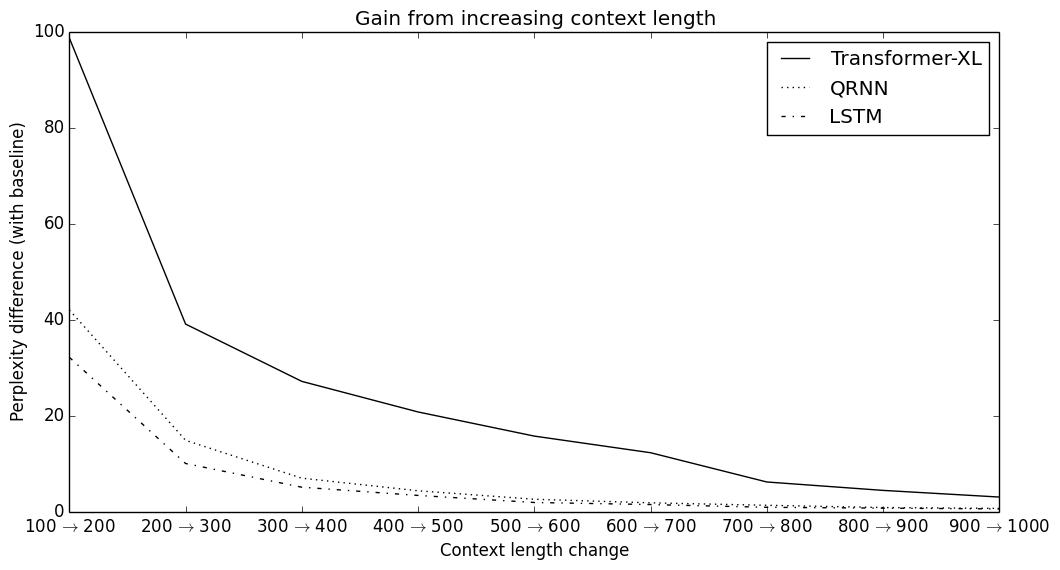}
		\caption{Transformer-XL vs RNNs}
		\label{fig:vsrnn}
	\end{subfigure}
	\begin{subfigure}[b]{0.5\textwidth}
		\includegraphics[width=\textwidth]{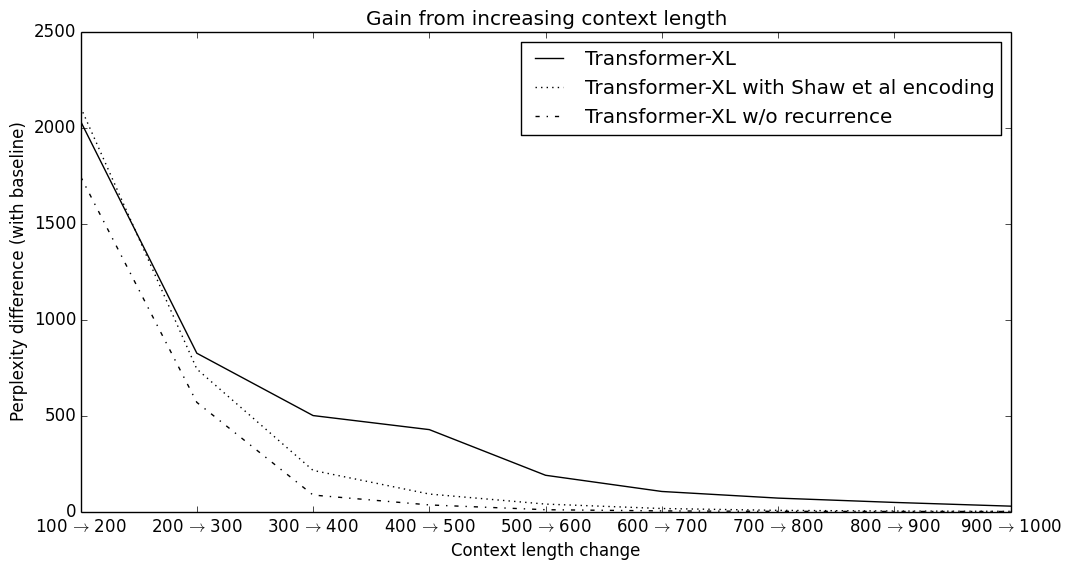}
		\caption{Transformer-XL vs Baseline}
		\label{fig:vsbase}
	\end{subfigure}
	\caption{Visualizing unnormalized relative perplexity gains with $r = 0.1$.}
	\label{fig:gain}
\end{figure}

\begin{figure}[!h]
	\begin{subfigure}[b]{0.5\textwidth}
		\includegraphics[width=\textwidth]{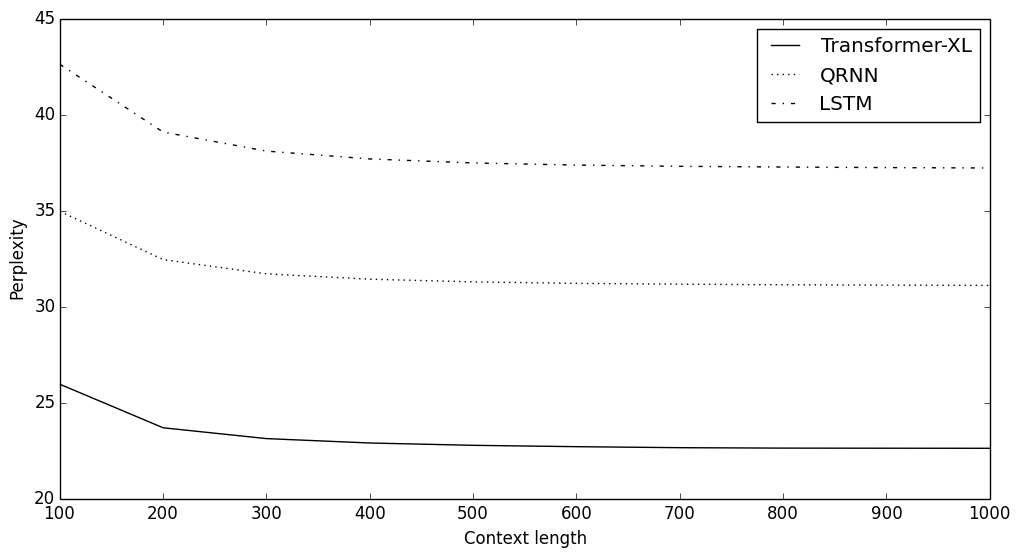}
		\caption{Transformer-XL vs RNNs}
		\label{fig:vsrnn}
	\end{subfigure}
	\begin{subfigure}[b]{0.5\textwidth}
		\includegraphics[width=\textwidth]{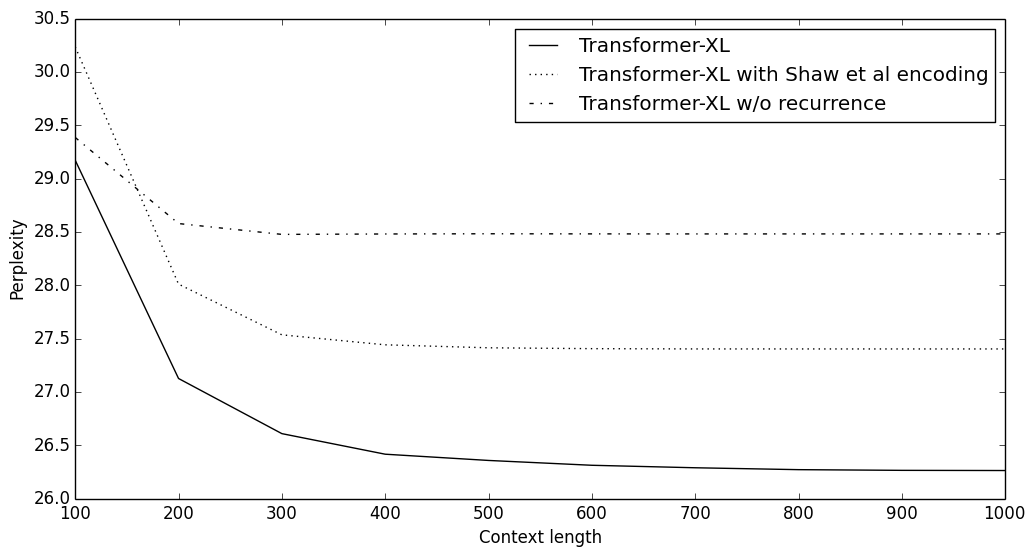}
		\caption{Transformer-XL vs Baseline}
		\label{fig:vsbase}
	\end{subfigure}
	\caption{Perplexity vs context length.}
	\label{fig:context}
\end{figure}

In this section, we describe the details of the metric RECL. Let $\mathcal{M} = \{m_1, m_2, \cdots, m_N\}$ be a model group consisting of $N$ models. Let $l_i(c, t)$ denote the loss of model $m_i$ on the $t$-th token in the corpus with a context length $c$. Concretely, the loss can be written as
\[
l_i(c, t) = - \log P_{m_i}(x_t | x_{t - 1}, \cdots, x_{t - c})
\]
where $P_{m_i}$ is the probability distribution given by model $m_i$, and $x_t$ is the $t$-th token in the corpus. Given a short context length $c$ and a long context length $c'$ such that $c' \geq c$, we can further define a baseline for each position $t$,
\[b(c, t) = \min_{i = 1}^N l_i(c, t)\]

The \textit{relative loss} of $m_i$ w.r.t. the model group $\mathcal{M}$ is written as
\[
f_i(c, c') = \frac{1}{|\mathcal{T}|} \sum_{t \in \mathcal{T}} \min \left( b(c, t), l_i(c', t) \right)
\]
The above equation uses the minimum loss of all models on the short length $c$ as a baseline, and only losses smaller than the baseline will be effectively counted towards the relative loss. This enables fair comparison between multiple models because all models with a long context length $c'$ need to improve over the same baseline. Sometimes we only care about those positions where the baseline performs poorly (which means short-term dependency with context length $c$ is not sufficient), so given a ratio parameter $r$, we define the set $\mathcal{T}$ is the above equation as
\[\mathcal{T} = \text{top-}r \text{~positions~} t \text{~with largest~}b(c, t)\]

The \textit{relative gain} is subsequently defined as the relative perplexity reduction:
\[
g_i(c, c') = \frac{\exp f_i(c, c) - \exp f_i(c, c')}{\exp f_i(c, c)}
\]

Given a step size $\Delta$, we then use an algorithm to find the RECL by thresholding the relative gain:
\begin{enumerate}
	\item Set initial short context length $c$, and long context length $c' = c + \Delta$
	\item Compute $g_i(c, c')$. If $g_i(c, c') < 0.01$, return $\text{RECL} = c$. If $g_i(c, c') \geq 0.01$, set $c = c', c' = c + \Delta$ and go to step 1.
\end{enumerate}

In Figure \ref{fig:gain}, we visualize the unnormalized relative perplexity gains $(\exp f_i(c, c) - \exp f_i(c, c'))$ with various pairs of $(c, c')$ when $r = 0.1$. It is clear that Transformer-XL has a longer RECL compared to RNNs and other baselines because the relative gains are substantially larger.

For reference, we plot the perplexities with varying context lengths in Figure \ref{fig:context}. The y-axis denotes the ``normal'' perplexity (not calibrated by baselines).

\section{Attention Visualization}
In this section, we provide some visualization of the attention learned by the SoTA model on the WikiText-103 validation set.
Recall that, this model has 16 10-head transformer layers and relies on a memory of length 640.

\begin{figure}[!h]
	\includegraphics[width=\linewidth]{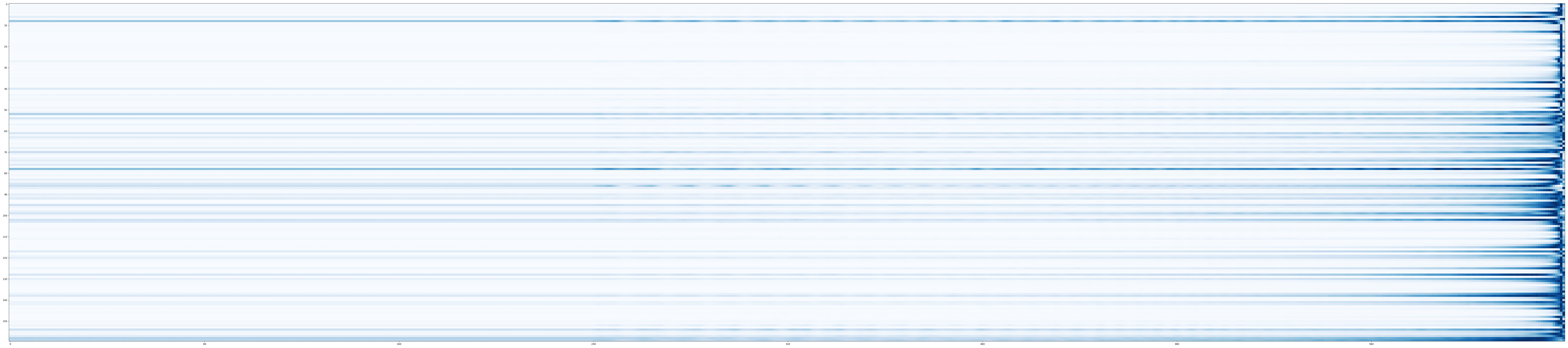}
	\caption{Average attention over the previous 640 tokens, where each row corresponds to a attention head and each column corresponds to a relative location. There are totally 160 attention heads, and every 10 heads come from a single layer. Darker colors indicate higher values.}
	\label{fig:visattn-global}
\end{figure}
The first visualization aims at revealing the overall trend of where the model is attending.
Specifically, for each attention head of each layer, we average the attention distributions of all tokens in the validation set.
This is shown in Fig. \ref{fig:visattn-global}. As we can see, the overall trend is to focus more on the nearby tokens than the faraway ones.
However, it is also very clear that some attention heads have a wider attention distribution over the entire memory span, notably the head 8 from layer 1, head 78 from layer 8, and the head 158 from layer 16.

\begin{figure}[!h]
	\begin{subfigure}[b]{\linewidth}
		\includegraphics[width=\textwidth]{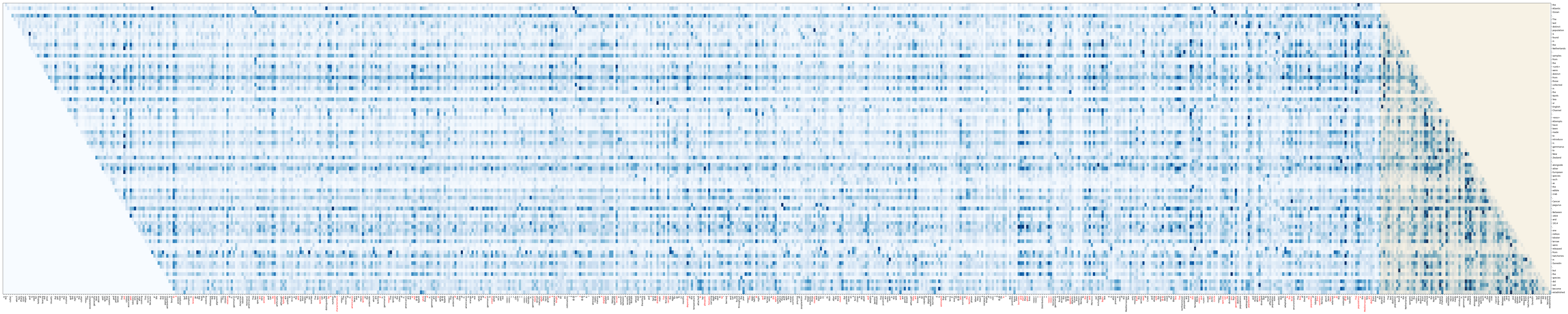}
		\caption{Head 8 from layer 1.}
		\label{fig:visattn-8}
	\end{subfigure}
	\begin{subfigure}[b]{\linewidth}
		\includegraphics[width=\textwidth]{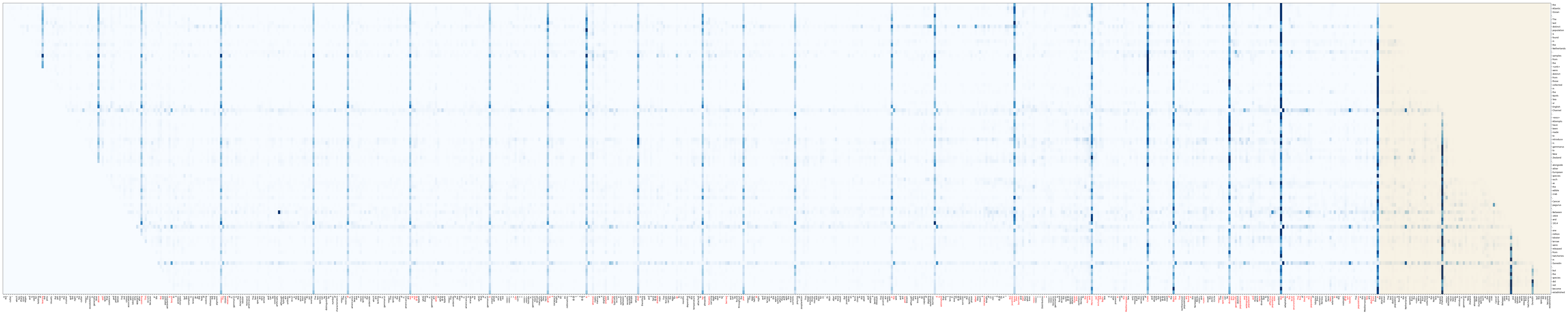}
		\caption{Head 78 from layer 8.}
		\label{fig:visattn-78}
	\end{subfigure}
	\begin{subfigure}[b]{\linewidth}
		\includegraphics[width=\textwidth]{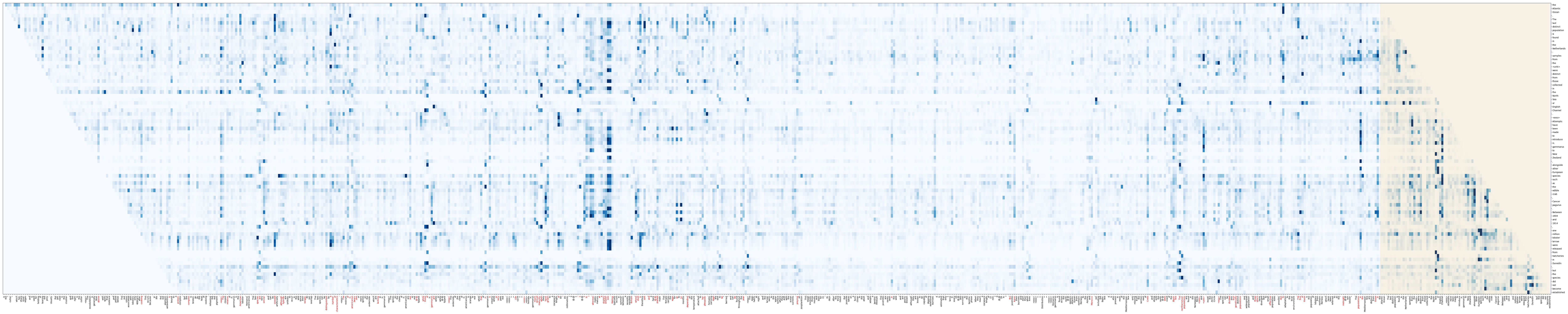}
		\caption{Head 158 from layer 16.}
		\label{fig:visattn-158}
	\end{subfigure}
	\caption{Visualization of the three heads with a wide attention range. Each row corresponds to a target location/token and each column corresponds to a context location/token. Tokens in the memory that have top 20\% attention values are highlighted in red.}
	\label{fig:visattn-pick}
\end{figure}
Since we are focused on learning long-range dependency, we are especially interested in these heads with a wider attention span.
Thus, in the second set of visualization, we pick the three notable heads mentioned above, and visualize their attention behavior for a randomly chosen position, as shown in Fig. \ref{fig:visattn-pick}.
Here, we see three different patterns of wider attention:
\begin{itemize}[leftmargin=*]
	\item For the head 8 in the 1st layer, we see an almost uniform attention over the entire memory span. This is quite intuitive, as lower-level layers needs to screen the entire memory span to decide where to focus for higher-level layers
	\item For the head 78 in the 8th layer (a middle-level layer), we see a very sparse attention pattern scattered in all ranges of the memory. Again, this well fits our intuition that as information accumulates, the network may focus on some particular position with special interests.
	\item For the head 158 in the 16th layer (i.e. the last layer), each target location (corresponding to each row) has its own distinct sparse focus, differing from head 78 where target locations largely share the same attentive location in memory. Meanwhile, the pattern is also different from the case of head 8, where a few locations are clearly attended more than others.
\end{itemize}

\begin{figure}[!h]
	\begin{subfigure}[b]{\linewidth}
		\includegraphics[width=\textwidth]{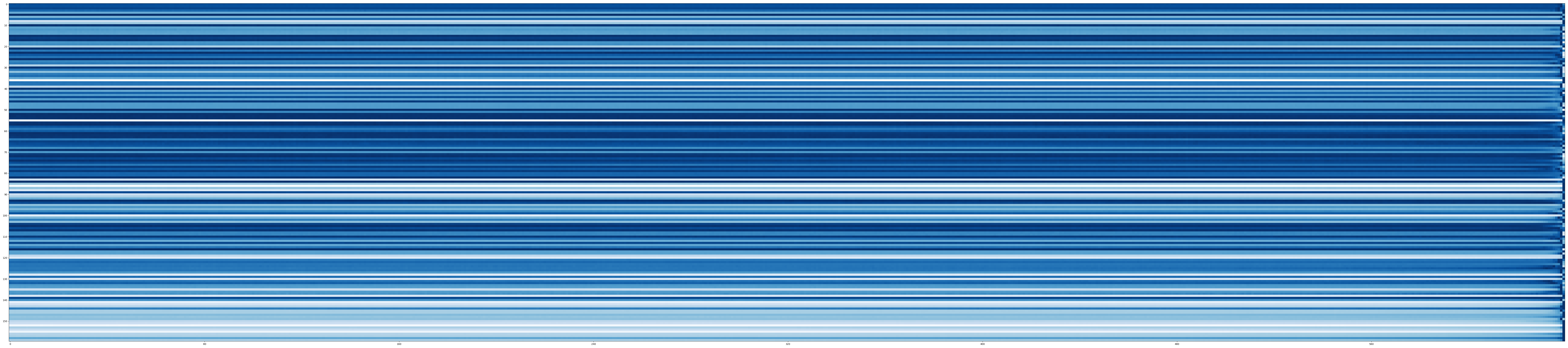}
		\caption{Term $(a)$.}
		\label{fig:visattn-A}
	\end{subfigure}
	\begin{subfigure}[b]{\linewidth}
		\includegraphics[width=\textwidth]{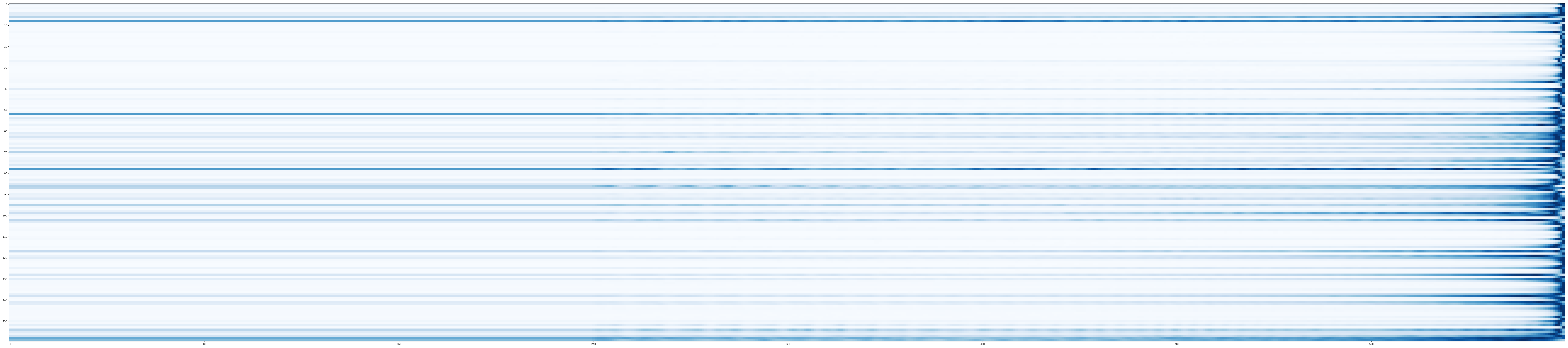}
		\caption{Term $(b)$.}
		\label{fig:visattn-B}
	\end{subfigure}
	\begin{subfigure}[b]{\linewidth}
		\includegraphics[width=\textwidth]{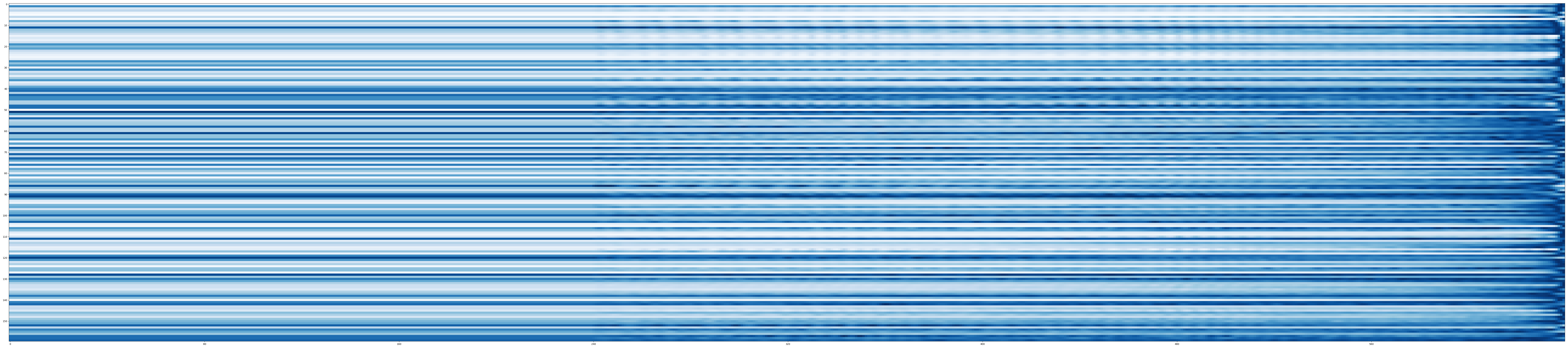}
		\caption{Term $(d)$.}
		\label{fig:visattn-D}
	\end{subfigure}
	\caption{Visualization of the three terms in computing the attention score. Each row corresponds to a attention head and each column corresponds to a relative location.}
	\label{fig:visattn-decomp}
\end{figure}
Finally, as we have discussed in section \ref{sec:rel-pos-embed}, the attention score can be decomposed into four intuitive terms.
Here, we want to further investigate how these four terms contribute to the overall attention trend in Fig. \ref{fig:visattn-global}.
Since the term $(c)$ represents the global content bias, i.e., the prior importance of each word regardless of the context, we will leave it out and focus on the terms $(a)$, $(b)$ and $(d)$.
So, for each term, we take the Softmax w.r.t. the memory span and average the resulted distribution of all tokens in the validation set.
The results are visualized in Fig. \ref{fig:visattn-decomp}:
\begin{itemize}[leftmargin=*]
	\item Since term $(a)$ is fully content-based addressing, when averaging over all target words, the result is essentially uniform over the entire context, except for a few very close words, which are likely to be semantically similar to the target word.
	\item The overall trend of term $(b)$ highly resembles that of the entire attention distribution in Fig. \ref{fig:visattn-global}. It suggests that the global trend of focusing on the nearby context is largely contributed by this content-dependent positional bias.
	\item The overall trend of term $(d)$ is also focusing more on nearby words. However, compared to the trend of term $(b)$, it is clearly flatter and biases towards a longer context.
\end{itemize}

\section{Generated Text} \label{sec:gen}
In this section, we present some generated text from our best model trained the Wikitext-103 dataset.
We seed the our Transformer-XL with a context of at most 512 consecutive tokens randomly sampled from the test set of Wikitext-103. Then, we run Transformer-XL to generate a \textit{pre-defined} number of tokens (500 or 1,000 in our case). For each generation step, we first find the top-40 probabilities of the next-step distribution and sample from top-40 tokens based on the re-normalized distribution.
To help reading, we detokenize the context, the generated text and the reference text. Three generated examples are shown in Tables \ref{tab:gen-1}, \ref{tab:gen-2}, and \ref{tab:gen-3}. Note that we do not perform any cherry picking and present the first three examples we generate in the paper.
In the text, ``= text ='', ``= = text = ='' and ``= = = text = = ='' denote the Wikipedia page tile, section title and subsection title, respectively, due to the original data preprocessing procedure of Wikitext-103~\cite{merity2016pointer}. 

%
%

As we can see, though only trained on 100M tokens, Transformer-XL is a strong model at generating long text articles, particularly in the following aspects:
\begin{itemize}[topsep=0em,itemsep=0em,parsep=0.2em]
	\item Transformer-XL is able to structurally maintain the sectional arrangement of Wikipedia.
	\item Transformer-XL manages to semantically stay on the same topic throughout the course of generation.
	\item Long-range references are common in the generated text.
	\item Transformer-XL often generates novel content that is not present in the training data.
\end{itemize}
For more detailed explanation of the interesting observations in each example, please refer to the corresponding caption.

Despite the overall excellence of the generation quality, the model can only perceive the seed context and hallucinate what to generate based on the limited knowledge (100M tokens only) it is trained on. As a result, the generated text sometimes looks clearly relevant but not close enough or to the point compared to what human writer would do. 
That said, we believe this issue is mostly a problem of limited training data size and could be alleviated by using a larger training set.

\begin{table}[!h]
\centering
\scriptsize
\begin{tabular}{p{7.8cm}|p{7.8cm}}
	\toprule
	\multicolumn{2}{p{16cm}}{\textbf{Context:}} \\
	\multicolumn{2}{p{16cm}}{Kershaw started the 2010 season by posting a 3.07 ERA in April, but did so by walking 22 batters in 29 innings. On May 4, he had his worst start of his career against the Milwaukee Brewers at Dodger Stadium, throwing just 57 pitches in 11 / 3 innings, while retiring only four of the 13 batters he faced — including the pitcher. He was booed loudly upon being pulled from the game. Kershaw said after the game, " I didn't give our team any kind of chance. It's just not a good feeling to let your teammates down, let everybody down. It stings, it hurts. I 've got to figure things out. " 
Kershaw rebounded his next start by pitching an 8 inning two-hitter and out-dueling the then undefeated Ubaldo Jim\'{e}nez. He credited his control of the slider being the major turning point for him. Later in the season, he was suspended for five games after hitting Aaron Rowand of the Giants with a pitch in a game on July 20. The incident occurred after both teams were given a warning following Giants ace Tim Lincecum hitting Matt Kemp earlier in the game. He threw his first career complete game shutout on September 14, 2010 also against San Francisco and finished the season with a record of 13 – 10 and a 2.91 ERA in 32 starts, pitching 2041 / 3 innings and recording 212 strikeouts. 

= = = = 2011 season: 1st Cy Young Award = = = = 

After finishing the 2010 season strong, the Dodgers named Kershaw as the Opening Day Starter for the 2011 season. On May 29, he pitched the second complete-game shutout of his career, striking out 10 while winning a two-hitter against the Florida Marlins, 8 – 0; he also had two singles and an RBI, scoring twice in the game. He produced his third career shutout on June 20, a two-hit, 11-strikeout effort against the Detroit Tigers. Kershaw became the first Dodgers starter to strike out the side in the 9th inning since Sandy Koufax's perfect game. In his next start, on June 26, Kershaw pitched another complete game (against the Los Angeles Angels of Anaheim ). He became the first Dodger starter to have back-to-back complete game victories since Jeff Weaver in the 2005 season and the first Dodger to have double-digit strikeouts in consecutive starts since Chan-Ho Park in the 2000 season. He was awarded the National League Player of the Week award for the week of June 20 – 26 as a result of those two starts. Midway through June, Kershaw had amassed 32 career victories } \\
	\midrule
	\textbf{XL Generation:} & \textbf{Reference:} \\	
	, tying him with Luis Castillo for the third best start total in the league and trailing only Johnny Bench and Eddie Mathews (37). The Dodgers signed Kershaw to a one-year deal on July 14, 2011, with a player option for 2012, in a deal worth just under \$ 300,000. 
The Dodgers entered the 2011 season as the major league leaders, and Kershaw had the best record in the National League in saves (39), games started (78) and walks (120); he also had the lowest win / loss ratio in the league (1.87), and he had the worst on-base percentage (.349). He had the best ERA (2.00 ), most wins (16 ), or most shutouts (29) in his career (25), while leading the league in losses per nine innings pitched (4.01). Following the 2011 season, the Dodgers traded Kershaw to the Kansas City Royals on July 29 for pitcher Javier Torres. 

= = = Kansas City Royals = = =

= = = = 2012 season = = = = 

During spring training, Kershaw played very well. He was selected to spring training as a relief pitcher for the Royals for the 2012 season. After an injury to closer Javier Vázquez, he was activated on April 29 to replace Matt Holliday in the Royals ' starting rotation. In his only start with the Royals, on August 6, 2012, Kershaw struck out five batters in seven innings pitched to help the Royals to their first victory in franchise history. On September 27, 2012, it appeared Kershaw was going to pitch a complete game shutout against the Detroit Tigers, but did not manage to do so since the Tigers won 3 – 1. At the conclusion of the season, Kershaw was named Major League Baseball's Most Valuable Player, was chosen to the All-Star Game at Busch Stadium and was named to the All-Star Game as the starting pitcher at shortstop. The Royals announced on February 4, 2013 that Kershaw would spend 2013 as starting pitcher, though he was expected to miss the season as a result of a shoulder injury. 

= = = = 2013 season = = = = 

On May 17, 2013, Kershaw sustained \underline{another back injury} and did not start in August and October 2013. He appeared in 22 starts, all starts, finishing with a strikeout-to-walk ratio of 1.50 and a 2.91 ERA. He also had the third most strikeouts in the league: 10. On May 20, 2013, he & , a 3.15 ERA and 593 career strikeouts in 568.2 innings. According to the Elias Sports Bureau, Kershaw was the first 23-year-old pitcher to have that many victories, an ERA that low and an average of more than one strikeout per inning since ERA became an official statistic in 1910. 
Kershaw was selected to the National League team for the 2011 Major League Baseball All-Star Game, his first All-Star selection. In the month of July, Kershaw was 4 – 1 with a 2.02 ERA and NL-leading 45 strikeouts, earning him the National League Pitcher of the Month Award. On August 23, he struck out Matt Holliday of the St. Louis Cardinals for his 200th strikeout of the season and became the 10th Dodger pitcher to record back-to-back 200 strikeout seasons and the first since Chan-Ho Park did it in the 2001 season. 
Kershaw finished the 2011 season by leading the NL with 21 wins, 248 strikeouts and a 2.28 ERA, winning the NL pitching Triple Crown, the first Triple Crown winner since Jake Peavy of the 2007 San Diego Padres and the first Dodger since Sandy Koufax won it in the 1966 season. Justin Verlander of the Detroit Tigers won the American League Triple Crown the same season, marking the first major-league season since 1924 to feature Triple Crown-winning pitchers in both leagues. Kershaw's 21 wins were the most by a Dodger pitcher since Orel Hershiser won 23 during the 1988 season. His ERA was the lowest by a Dodger since Hershiser's 2.03 in the 1985 season, his strikeouts were the most by a Dodger since Koufax's 317 in 1966 and his 233 1 / 3 innings pitched were the most since Chan Ho Park pitched 234 in 2001. Since 1965 when Koufax did it, Peavy and Kershaw are only two pitchers in the National League have led the league in wins, strikeouts, ERA, and WHIP (walks plus hits per inning pitched). Kershaw also became just the second <unk> to have a 240-plus strikeouts in a season before the age of 24, joining Vida Blue. 
After the season, Kershaw was awarded the Warren Spahn Award as the best left-handed pitcher in 2011, the Players Choice Award for Most Outstanding National League pitcher, the Gold Glove Award as the top fielding pitcher in the NL and the Sporting News (TSN) National League Pitcher of the Year. He was additionally selected as the starting pitcher for the TSN NL All-Star Team. On November 17, he was honored with the National League Cy Young Award, making him the youngest Cy Young winner since Dwight Gooden 
 \\
	\bottomrule
\end{tabular}
\captionsetup{singlelinecheck=off}
\caption[]{\small 
	Example 1 -- 500 tokens generated by XL using a snippet from the Wikitext-103 test set as initial context. The sample is randomly generated without any cherry picking. 
	\vspace{0.4em}
	
	Original Wikipedia page: {\footnotesize \url{https://en.wikipedia.org/wiki/Clayton_Kershaw}}
	\vspace{0.4em}
	
	There are many interesting observations from this example:
	\begin{itemize}[leftmargin=*,itemsep=0em,topsep=0.4em,parsep=0.4em]
		\item Firstly, Kershaw never went to Royals in real life. Despite that, Transformer-XL stays on the fully imagined topic and keeps hallucinating the experience of Kershaw in Royals across the generated text.
		\item Secondly, notice that XL correctly tracks the chronological order from 2011 to 2012 and to the finally 2013 season in the section titles. 
		\item  In addition, notice that Transformer-XL accurately uses the the phrase ``\underline{another back injury}'' in the 2013 season paragraph, since it has talked about one earlier injure in the 2012 season. This shows again Transformer-XL's ability of capturing long-term dependency.
	\end{itemize}
}
\label{tab:gen-1}
\end{table}

\begin{table}[!h]
\centering
\scriptsize
\begin{tabular}{p{7.8cm}|p{7.8cm}}
	\toprule
	\multicolumn{2}{p{16cm}}{\textbf{Context:}} \\
	\multicolumn{2}{p{16cm}}{= = Distribution = = 

Species range across the Neotropics from Mexico in the north to Bolivia, Paraguay, and southern Brazil in the south. According to <unk> and coauthors, three species are found in Mexico, four in Central America, and 62 in South America. Three species are present in the Caribbean — two in Trinidad and Tobago, along the southern edge of the region, and one in Haiti. 

= = Habitat and ecology = = 

<unk> includes both large trees and small acaulescent palms which occupy a number of different ecological niches. Dense stands of some of the larger species are conspicuous elements on the landscape, while smaller species are found in both in the forest understorey and in savannas. 
Disturbance has been implicated in the formation of vegetation dominated by large <unk> species. In seasonally dry Amazonian forests the density of large adult A. <unk> palms was correlated with canopy openness; the species also dominates savannas formed by repeated forest fires in Trinidad and Tobago. <unk> speciosa forms pure stands in many parts of Brazil where natural forest vegetation has been cleared. Similarly, stands of A. <unk> in Bahia, Brazil (which are cultivated for <unk> fibre) are managed using fire — the seedlings survive cutting and burning, and are able to dominate burned forest patches. 
The fruit are dispersed by animals; fruit which are not dispersed frequently suffer seed predation by <unk> beetles. Certain species of <unk> have been mentioned as examples of " anachronistic " species which are adapted for dispersal by now-extinct Pleistocene megafauna. On <unk> Island, <unk>, in the Brazilian Amazon, <unk> <unk> fruit were consumed by tapirs, collared peccaries, deer and primates. Rodents, including agoutis, fed upon the fruit and, as the fruit availability declined, they fed on the seeds. Other dispersers of <unk> fruit include Crested <unk> which consume the fruit and disperse the seeds of A. <unk> in the Brazilian Pantanal. 

= = Uses = = 

<unk> species have a long history of human utilisation. <unk> <unk> <unk> seeds have been found in archaeological sites in Colombia dating back to 9000 BP. A variety of species remain important sources of edible oil, thatch, edible seeds and fibre. The leaves of <unk> <unk> and A. <unk> are used extensively for thatching. Several species are oil palms, with A. speciosa among the most important economically. Products extracted from A. speciosa were reported to support over 300,000 households in the Brazilian state of Maranhão in 2005, and in 1985 it was estimated to support over 450,000 households throughout the Brazil. <unk> fibres, extracted from the leaf bases of A. <unk>, are commercially important, and generated about US \$} \\
	\midrule
	\textbf{XL Generation:} & \textbf{Reference:} \\	
	700,000 in 2013 on the black market. In 2004 the plant was reported to be responsible for a net profit of around US \$ 3 million in Brazil. A 2012 study found A. speciosa to cause agricultural production at least 2 – 3 times greater than those achieved without prior knowledge of land management.

\underline{= The Tale of Mrs. Tittlemouse =}

The Tale of Mrs. Tittlemouse is a children's book written and illustrated by H. M. Potter. It was first published in February 1910 by Frederick Warne \& Co. and included illustrations by Frederick Warne. Mrs. Tittlemouse was intended to be a children's book, rather than a musical, and the tale describes her attempts to find her own home. Mrs. Tittlemouse is an average kitten and the tale follows a mischievous rabbit named Mr. Tittlemouse who tries to kill her in an attempt to get a ride on a horse. Potter later wrote of the rabbit and her attempts to kill her, " If [ she ] were a fox, I should like to have been able to show her the way. " 
Potter's work was inspired by The Tale of Little Black Sambo and her two nieces, Petunia and Petunia. It was the first book in the Potter family to be published in a hardback format, and was also the first book in the Peter Rabbit series. The tale was released on 13 November 1910 and was accompanied by a greeting card by Potter's brother-in-law, Frederick Warne. 
The tale was well received by critics and children, and sold over 600,000 copies in its first few years. In the first year of its publication, Potter's The Tale of Mrs. Tittlemouse won children's book awards in England, America and Australia. It was selected as a " recommended reading " by children in the US and Canada. 

= = Plot summary = = 

The tale opens with the pet rabbit Mrs. Tittlemouse wandering through a forest in a small litter of four. He is shot and badly injured by a deer after she attempts to escape. Mrs. Tittlemouse decides that she wants to find her own home, because she is eager to go on her own. She goes alone to the farm where she makes a little money by selling a few seeds and building a small cabin in the woods. She is approached by a wealthy hunter named Mr. Tittlemouse, who tries to kill her but Mrs. Tittlemouse kills him by stuffing a rope into his nose and killing him. She is rescued by Mr. Tittlemouse's wife Ruth, but Mrs. Tittlemouse then leaves the woodland with the baby. When she is spotted by  & 20 million in annual income to Brazilian farmers in 1996.

\underline{= The Heart of Ezra Greer =}

The Heart of Ezra Greer is a 1917 American silent drama film produced by the Thanhouser Company and directed by Emile <unk>. The film focuses on Ezra Greer, a successful middle-aged man who searches for his college age daughter, Mary. The wayward Mary was romanced and abandoned by Jack <unk>, later bearing his child. Once Ezra becomes broke he finds employment as the valet for Jack <unk>. After Jack's engagement to a cabaret girl, Mary becomes upset and leaves her child at Jack's home. Contrary to Jack's wishes, Ezra keeps the child and Jack ultimately reveals that the child is his own. Ezra convinces Jack to make things right and Ezra convinces the cabaret girl to leave Jack. After a carriage accident in which the baby is injured, Ezra and Jack rush to the hospital and find Mary as a nurse crying over the child. The film ends with the marriage of Jack and Mary. The film was released by Pathé on October 7, 1917. The film was the final release from Thanhouser and was deemed to be an average film by most reviewers. Criticism for the film hinged on far-fetched coincidences to drive the plot. The film is presumed lost. 

= = Plot = = 

The film follows Ezra Greer, a middle-aged man who has worked hard since his youth. He cares deeply for his motherless daughter, Mary, but was unable to attend the annual commencement at her co-educational college. He awaits for her to return from college, but Mary leaves with her romantic interest, Jack <unk>. On promise of marriage and wealth, Mary is romanced and gives birth to a fatherless child. Without word from his daughter, Ezra resigns from his job and attempts to seek her out and finds a poor motherless child, Marie. With Ezra's money exhausted he seeks employment and finds it as the valet of Jack. 
One day, Mary seeks an announcement of Jack's engagement to a cabaret girl known as " The Baby Vamp ". Bitter over the prospect of her child's future, she leaves the child at Jack's home during his absence with a note. Jack orders Ezra to take the baby to an orphanage, but Marie begs Ezra to keep him. After continually seeing the child, Jack is overcome with remorse and explains to Ezra and seeks his advice. Not knowing he was making the case for his own daughter, Ezra convinces Jack to seek out Mary and forget the Baby Vamp. The Baby  \\
	\bottomrule
\end{tabular}
\captionsetup{singlelinecheck=off}
\caption[]{\small 
	Example 2 -- 500 tokens generated by XL using a snippet from the Wikitext-103 test set as initial context. The sample is randomly generated without any cherry picking. 
	\vspace{0.4em}
	
	Original Wikipedia page: {\footnotesize \url{https://en.wikipedia.org/wiki/The_Tale_of_Mrs._Tittlemouse}}. \vspace{0.4em}
	
	This example exhibit some additional interesting properties of Transformer-XL: 
	\begin{itemize}[leftmargin=*,itemsep=0em,topsep=0.4em,parsep=0.4em]
		\item After finishing the last paragraph of the seed context, both the reference and generated text start a new topic (i.e., Wikipedia page), as marked by the single ``\underline{= title =}'' line. 
		This suggests the model has the ability of identifying the end of a topic / page, and randomly starting with a new topic.
		\item Even more interestingly, a newly-started page is on a book called ``The Tale of Mrs. Tittlemouse''. Transformer-XL manages to copy the same book title and some related information from the training set, but hallucinates \textit{novel} content of the book. 
		This demonstrates a degree of generalization instead of memorization. Please refer to the original book content at the Wikipedia page.
	\end{itemize}
}
\label{tab:gen-2}
\end{table}

\clearpage

\begin{center}
\scriptsize
\begin{longtable}{p{7.8cm}|p{7.8cm}}
	\toprule
	\multicolumn{2}{p{16cm}}{\textbf{Context:}} \\
	\multicolumn{2}{p{16cm}}{
= Battle of D\"{u}renstein = 

The Battle of D\"{u}renstein (also known as the Battle of <unk>, Battle of <unk> and Battle of <unk>; German: <unk> bei <unk> ), on 11 November 1805 was an engagement in the Napoleonic Wars during the War of the Third Coalition. Dürenstein (modern <unk>) is located in the <unk> Valley, on the River Danube, 73 kilometers (45 mi) upstream from Vienna, Austria. The river makes a crescent-shaped curve between <unk> and nearby Krems an der Donau and the battle was fought in the flood plain between the river and the mountains. 
At Dürenstein a combined force of Russian and Austrian troops trapped a French division commanded by Théodore Maxime Gazan. The French division was part of the newly created VIII Corps, the so-called Corps Mortier, under command of \'{E}douard Mortier. In pursuing the Austrian retreat from Bavaria, Mortier had over-extended his three divisions along the north bank of the Danube. Mikhail <unk> Kutuzov, commander of the Coalition force, enticed Mortier to send Gazan's division into a trap and French troops were caught in a valley between two Russian columns. They were rescued by the timely arrival of a second division, under command of Pierre Dupont de l 'Étang. The battle extended well into the night. Both sides claimed victory. The French lost more than a third of their participants, and Gazan's division experienced over 40 percent losses. The Austrians and Russians also had heavy <unk> to 16 <unk> perhaps the most significant was the death in action of Johann Heinrich von Schmitt, one of Austria's most capable chiefs of staff. 
The battle was fought three weeks after the Austrian capitulation at Ulm and three weeks before the Russo-Austrian defeat at the Battle of Austerlitz. After Austerlitz Austria withdrew from the war. The French demanded a high indemnity and Francis II abdicated as Holy Roman Emperor, releasing the German states from their allegiance to the Holy Roman Empire. 

= = Background = = 

In a series of conflicts from 1803-15 known as the Napoleonic Wars, various European powers formed five coalitions against the First French Empire. Like the wars sparked by the French Revolution (1789 ), these further revolutionized the formation, organization and training of European armies and led to an unprecedented militarization, mainly due to mass conscription. Under the leadership of Napoleon, French power rose quickly as the Grande Armée conquered most of Europe, and collapsed rapidly 
} \\
	\midrule
	\textbf{XL Generation:} & \textbf{Reference:} \\	
	after the French conquest of Italy. This victory marked the beginning of the Second Coalition. Napoleon's rapid advance caused Austria – Russia, Britain and Russia to make peace negotiations. The Russian army under Prince Mikhail Mikhailovich Mikhailovich Chaffee, commander of the Imperial Russian Army in Switzerland, was able to drive off the Austrians in the Battle of Stockach (1805) and to take Pressburg (modern \"{O}3 Austria) from the Austrians. At an early stage the Austro-Hungarian fleet had been damaged in a storm off Cape Matapan; this allowed the Allies to launch a full-scale invasion of Italy in February 1805. The Battle of Austerlitz was the decisive French victory against Napoleon and the largest naval battle in history, involving more modern-day European navies. 
The French military situation worsened as Napoleon faced several threats related to his newly formed Army of Europe, particularly Russia itself, which was now on the verge of collapse. The Russian general and friend of Napoleon, Alexander, had been dissatisfied with the conduct of the Austrians; he was still in conflict with Emperor Napoleon, the French Republic's king, who had declared war on Austria during the summer of 1804. With the war over, Napoleon decided to launch a second invasion of Italy in October 1805. 

= = Prelude = = 

\magenta{In July 1805}, the French 1st Army entered southern Italy. The army, under the command of Marshal Marmont, were reinforced by a few battalions of infantry under Claude General Auguste de Marmont at the town of Philippsburg and another battalion at Belluno. On \magenta{17 September 1805}, the army marched from Belluno towards Krems. By \magenta{29 September}, they had reached Belluno and conducted its advance against a small Austrian force. By \magenta{31 September}, the whole force had been reinforced by a brigade from the Army of Tyrol under the command of Pierre Augereau. 
The Austrians were now under the command of Marshal Jean Victor Marie Moreau, a member of the Directory. Moreau had taken command of the Austrian invasion force in the spring of 1805. His command included the VI Corps commanded by Jean Baptiste Drouet de Ney and the VI Corps commanded by Generals Jean Victor Marie Moreau and Joseph Souham. Ney's corps consisted of the III. Corps and VI. Corps, which consisted of the III Corps and VI. Corps, located in the Austrian Netherlands, was commanded by Friedrich Joseph, Count Baillet de Latour. Moreau's army consisted of six divisions and several associated brigades. 

= = Aftermath = =

= = = First Coalition forces = = = 

On \magenta{9 October 1805} the French Army of the Danube was attacked by an Austrian army under Archduke Charles at the Battle of Austerlitz. Although Charles and Charles had not had much time to regroup, on \magenta{10 October}, he launched his attack on the Polish forces under Friedrich Joseph, Count of Lauenburg. After three days, Charles' army captured Lauenburg. The French forces pursued the Austrians to the Silesian border, where they encountered strong Austrian resistance. These conflicts forced the Austrians to retreat into Tyrol and Austria agreed to a truce. 
The Austrian army, commanded by Wenzel Anton Karl, Count of Merveldt, was reduced to around 10,000 men. It was initially planned that Archduke Charles would launch a counter-attack against the French army on the same day, as Napoleon had hoped, but this was not carried out. On \magenta{25 October}, Merveldt left Styria for Tyrol. On the same day, Austria launched its new offensive against the French at Ulm. Charles withdrew his army from the region for a third time at the Battle of Elchingen, under the overall command of the Austrian generals, Ferdinand and Friedrich Wilhelm of J\"{u}lich-Cleves-Berg. To prevent Archduke Charles from escaping from the battlefield, the commander of the Habsburg army, Archduke Charles, planned to occupy the fortress Linz; instead, he decided to force Franz von Hipper to surrender the city. However, as Charles moved to the south, Moreau arrived on the scene with additional soldiers – including the entire Imperial Guard – and defeated the Austrians at the Battle of Hohenlinden on \magenta{28 October}. 
The loss of Linz resulted in Austria's complete defeat at Hohenlinden. In the meantime, the French Army of Observation and Preparedness was reorganized into the Army of the Danube under Feldzeugmeister (Colonel-General) Friedrich Freiherr von Hotze. The army was composed of the I, IV, VI, VI, VII, VIII and IX Corps. With reinforcements from Italy and France, it formed new battalions, companies, and squadrons in the Austrian army. On \magenta{17 November 1804}, at the Battle of Jena-Auerstadt the Army of Silesia and the Army of Silesia joined forces, but by the time that the 
 & after the disastrous invasion of Russia in 1812. Napoleon's empire ultimately suffered complete military defeat in the 1813 – 14 campaigns, resulting in the restoration of the Bourbon monarchy in France. Although Napoleon made a spectacular return in 1815, known as the Hundred Days, his defeat at the Battle of Waterloo, the pursuit of his army and himself, his abdication and banishment to the Island of Saint Helena concluded the Napoleonic Wars. 

= = Danube campaign = = 

From 1803-06 the Third Coalition fought the First French Empire and its client states (see table at right ). Although several naval battles determined control of the seas, the outcome of the war was decided on the continent, predominantly in two major land operations in the Danube valley: the Ulm campaign in the upper Danube and the Vienna campaign, in the middle Danube valley. 
Political conflicts in Vienna delayed Austria's entry into the Third Coalition until 1805. After hostilities of the War of the Second Coalition ended in 1801, Archduke <unk> emperor's <unk> advantage of the subsequent years of peace to develop a military restructuring plan. He carefully put this plan into effect beginning in 1803 – 04, but implementation was incomplete in 1805 when Karl Mack, Lieutenant Field Marshal and Quartermaster-General of the Army, implemented his own restructuring. Mack bypassed Charles ' methodical approach. Occurring in the field, Mack's plan also undermined the overall command and organizational structure. Regardless, Mack sent an enthusiastic report to Vienna on the military's readiness. Furthermore, after misreading Napoleon's maneuvers in W\"{u}rttemberg, Mack also reported to Vienna on the weakness of French dispositions. His reports convinced the war party advising the emperor, Francis II, to enter the conflict against France, despite Charles ' own advice to the contrary. Responding to the report and rampant anti-French fever in Vienna, Francis dismissed Charles from his post as generalissimo and appointed his <unk> brother-in-law, Archduke Ferdinand, as commander. 
The inexperienced Ferdinand was a poor choice of replacement for the capable Charles, having neither maturity nor aptitude for the assignment. Although Ferdinand retained nominal command, day-to-day decisions were placed in the hands of Mack, equally ill-suited for such an important assignment. When Mack was wounded early in the campaign, he was unable to take full charge of the army. Consequently, command further devolved to Lieutenant Field Marshal Karl Philipp, Prince of Schwarzenberg, an able cavalry officer but inexperienced in the command of such a large army. 

= = = Road to Ulm = = = 

The campaign in the upper Danube valley began in October, with several clashes in Swabia. Near the Bavarian town of Wertingen, 40 kilometers (25 mi) northwest of Augsburg, on 8 October the 1st Regiment of dragoons, part of Murat's Reserve Cavalry Corps, and grenadiers of Lannes ' V Corps surprised an Austrian force half its size. The Austrians were arrayed in a line and unable to form their defensive squares quickly enough to protect themselves from the 4,000 dragoons and 8,000 grenadiers. Nearly 3,000 Austrians were captured and over 400 were killed or wounded. A day later, at another small town, <unk> south of the Danube <unk> French 59th Regiment of the Line stormed a bridge over the Danube and, humiliatingly, chased two large Austrian columns toward Ulm. 
The campaign was not entirely bad news for Vienna. At Haslach, Johann von Klenau arranged his 25,000 infantry and cavalry in a prime defensive position and, on 11 October, the overly confident General of Division Pierre Dupont de l'\'{E}tang attacked Klenau's force with fewer than 8,000 men. The French lost 1,500 men killed and wounded. Aside from taking the Imperial Eagles and <unk> of the 15th and 17th Dragoons, Klenau's force also captured 900 men, 11 guns and 18 ammunition wagons. 
Klenau's victory was a singular success. On 14 October Mack sent two columns out of Ulm in preparation for a breakout to the north: one under Johann Sigismund Riesch headed toward Elchingen to secure the bridge there, and the other under Franz von Werneck went north with most of the heavy artillery. Recognizing the opportunity, Marshal Michel Ney hurried the rest of his VI Corps forward to re-establish contact with Dupont, who was still north of the Danube. In a two-pronged attack Ney sent one division to the south of Elchingen on the right bank of the Danube. This division began the assault at Elchingen. At the same time another division crossed the river to the east and moved west against Riesch's position. After clearing Austrian pickets from a bridge, the French attacked and captured a strategically located abbey at 
 \\
	French approached Vienna, the Prussians had already surrendered. As the Austrians did not want to allow the war to continue, they decided to abandon their territories in the north and move their army to the north and west, cutting off Charles from Vienna. The Battle of Warsaw was fought on \magenta{23 November 1805} between the French Army of the Danube and the Austrian Army of Styria in the vicinity of Warsaw and Pressburg (modern Trnava, Slovakia). At that time Habsburg forces  & the top of the hill at bayonet point. The Austrian cavalry unsuccessfully tried to fend off the French, but the Austrian infantry broke and ran. In this engagement alone, the Austrians lost more than half their reserve artillery park, 6,000 (out of 8,000 total participants) dead, wounded or captured and four colors. Reisch's column also failed to destroy the bridges across the Danube. 
Napoleon's lightning campaign exposed the Austrian indecisive command structure and poor supply apparatus. Mack  \\
	\bottomrule
\captionsetup{singlelinecheck=off}
\caption[]{\small 
	Example 3 -- 1,000 tokens generated by XL using a snippet from the Wikitext-103 test set as initial context. The sample is randomly generated without any cherry picking.
	\vspace{0.4em} 
	
	Original Wikipedia page: {\footnotesize \url{https://en.wikipedia.org/wiki/Battle_of_D\%C3\%BCrenstein}}. 
	\vspace{0.4em}
	
	\begin{itemize}[leftmargin=*,itemsep=0em,topsep=0.4em,parsep=0.4em]
		\item Although this example is significantly longer, we can see that Transformer-XL is still able to stay on the same topic and makes up non-existing stories about the Napoleon wars.
		\item Notably, from the second section on, the generated text correctly follows a fine-grained chronological order \textit{on the level of month and day} to narrate events in 1805, except a mistake (1804 instead of 1805) near the end of the paragraph. To ease reading which we have highlighted all the date related phrases by \magenta{magenta} in the generation.
	\end{itemize}
}
\label{tab:gen-3}
\end{longtable}
\end{center}


\FloatBarrier

\end{document}